\pdfoutput=1

\documentclass[10pt, journal,twoside]{IEEEtran}

\ifCLASSINFOpdf
\else
\fi

\usepackage{cite}
\usepackage{amsmath}
\usepackage{amsfonts}
\usepackage{algorithm}
\usepackage{algorithmic}
\usepackage{threeparttable}
\usepackage{graphicx}
\usepackage{subfigure}
\usepackage{color}
\usepackage{float}
\usepackage{booktabs}
\usepackage{makecell}
\usepackage{multirow}
\usepackage{pifont}
\usepackage{amssymb}
\usepackage{mathrsfs}
\usepackage{tablefootnote}
\begin{document}

%
\title{An Empirical Study of Remote Sensing Pretraining}
%
%
%

\author{Di Wang,
        Jing Zhang,
        Bo Du,~\IEEEmembership{Senior Member,~IEEE,}
        Gui-Song Xia,~\IEEEmembership{Senior Member,~IEEE,}\\
        and Dacheng Tao,~\IEEEmembership{Fellow,~IEEE}
        
\thanks{D. Wang, B. Du and Gui-Song Xia are with the School of Computer Science, Wuhan University, Wuhan 430072, China (e-mail: wd74108520@gmail.com; dubo@whu.edu.cn; guisong.xia@whu.edu.cn). B. Du is the corresponding author.}
\thanks{J. Zhang is with the School of Computer Science, 
Faculty of Engineering, The University of Sydney, Australia (jing.zhang1@sydney.edu.au).}
\thanks{D. Tao is with the JD Explore Academy, China and is also with the School of Computer Science, Faculty of Engineering, The University of Sydney, Australia (dacheng.tao@gmail.com).}
}

%
%

\markboth{Journal of \LaTeX\ Class Files,~Vol.~14, No.~8, August~2015}{Wang \MakeLowercase{\textit{et al.}}: EMPIRICAL STUDY OF REMOTE SENSING PRETRAINING}
%



\maketitle

\begin{abstract}

  Deep learning has largely reshaped remote sensing (RS) research for aerial image understanding and made a great success. Nevertheless, most of the existing deep models are initialized with the ImageNet pretrained weights. Since natural images inevitably present a large domain gap relative to aerial images, probably limiting the finetuning performance on downstream aerial scene tasks. This issue motivates us to conduct an empirical study of remote sensing pretraining (RSP) on aerial images. To this end, we train different networks from scratch with the help of the largest RS scene recognition dataset up to now --- MillionAID, to obtain a series of RS pretrained backbones, including both convolutional neural networks (CNN) and vision transformers such as Swin and ViTAE, which have shown promising performance on computer vision tasks. Then, we investigate the impact of RSP on representative downstream tasks including scene recognition, semantic segmentation, object detection, and change detection using these CNN and vision transformer backbones. Empirical study shows that RSP can help deliver distinctive performances in scene recognition tasks and in perceiving RS related semantics such as ``Bridge'' and ``Airplane''. We also find that, although RSP mitigates the data discrepancies of traditional ImageNet pretraining on RS images, it may still suffer from task discrepancies, where downstream tasks require different representations from scene recognition tasks. These findings call for further research efforts on both large-scale pretraining datasets and effective pretraining methods. The codes and pretrained models will be released at https://github.com/ViTAE-Transformer/RSP.

\end{abstract}


%
\IEEEpeerreviewmaketitle

\section{Introduction}
%
%
%
%

\IEEEPARstart{W}ITH the development of geoinformatics technology, the earth observation fields have witnessed significant progress, where various remote sensing (RS) sensors and devices have been widely used. Among them, with the advantages of real-time, abundant amount, and easy access, the aerial image has become one of the most important data sources in earth vision to serve the requirements of a series of practical tasks, such as precision agriculture \cite{agriculture_1,zhang2020empowering} and environmental monitoring \cite{env_monit_1}. In these applications, aerial scene recognition is a fundamental and active research topic over the past years. However, because of the own characteristics of aerial images, it is still challenging to efficiently understand the aerial scene.

\begin{figure}[t]
  \centering
  \includegraphics[width=1\linewidth]{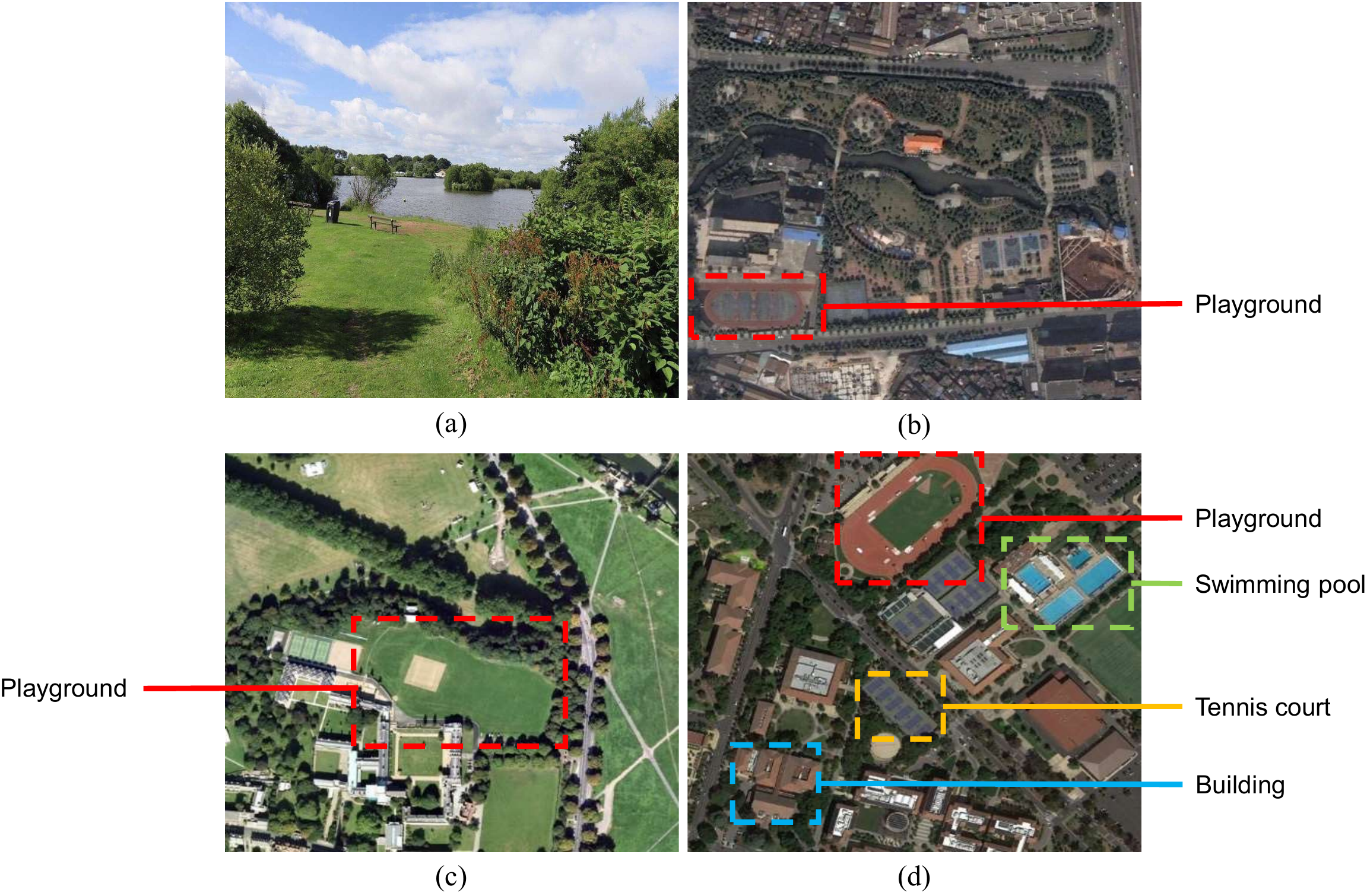}\\
  \caption{
  The challenges of aerial scene recognition. (a) and (b) are the natural image and aerial image belonging to the ``park'' category. (c) and (d) are two aerial images from the ``school'' category. Despite the distinct view difference between (a) and (b), (b) contains the playground that is unusual in the park scenes but usually exists in the school scenes like (d). On the other hand, (c) and (d) show different colors as well as significantly different spatial distributions of land objects like the playground and swimming pool. Here, (a) is obtained from http://travel.qunar.com/p-oi24486013-townhill\_country\_park by searching ``park'' on internet, while (b), (c), and (d) are the aerial images from the AID dataset.
  }
  \label{aerialscene}
\end{figure}

The aerial images are usually obtained by a camera in a birdview perspective lying on the planes or satellites, perceiving a large scope of land uses and land covers. The obtained aerial scene is usually difficult to be interpreted since the interference of the scene-irrelevant regions and the complicated spatial distribution of land objects. Specifically, it causes the issue of inter-class similarity in aerial scene understanding, i.e., some different scenes present similar characteristics, as well as the issue of large intra-class variations, where the scenes in the same category have discrepancies, as shown in Figure \ref{aerialscene}.

To tackle the above problems, it is necessary to obtain discriminative feature representations for different categories of aerial scenes. According to the difference in feature extraction methods, they can be divided into three types, i.e., the handcrafted features, the unsupervised learning features, and the supervised deep learning (DL) features. Initially, researchers directly utilize simple properties, such as  color \cite{asr_color}, texture \cite{asr_texture}, contour \cite{asr_shape}, spectral or their combination \cite{asr_spectral_comb1} to recognize different aerial scenes. Besides these intuitive attributes, there are also some well-designed feature descriptors. For instance, the scale-invariant feature transformation and histogram of oriented gradients. These handcrafted features usually perform well in simple scenes while being ineffective in complex scenes. They are usually regarded as shallow features from a modern view in the DL era, while interpreting complex scenes requires more semantic information, which can not be efficiently extracted by shallow-layer methods \cite{representlearning}. Compared with the above approaches, the unsupervised learning methods provide a feasible way to automatically extract the suitable features by adaptively learning the mapping functions or filters based on a set of handcrafted features or the raw pixel intensity values. The typical unsupervised learning methods include potential latent semantic analysis \cite{plsa} and bag-of-visual-words \cite{bovw}. Some simple feature enhancement methods such as the principal component analysis also belong to this category. Nonetheless, the encoded unsupervised features still have limited performance since no category supervision is explicitly used, which is useful for feature discrimination.

In recent years, with the superiority of automatically extracting deep features that reflect the inherent properties of objects, DL has achieved an impressive breakthrough in the computer vision (CV) field \cite{vgg,resnet,swint,xu2021vitae,upernet,zj_segmentation2,zj_segmentation3}, as well as the RS field \cite{assmn,zj_segmentation1,fullycontnet}. In the aerial scene recognition domain, the most commonly used deep models are the convolutional neural networks (CNN), which have a good ability for local perception and global perception, where the former is achieved via applying sliding window convolutions on the input image to extract the local features while the latter is achieved by stacking multiple convolutional layers to increase the receptive field. According to the training scheme, the ways of using CNN in aerial scene recognition methods can be divided into three categories, i.e., training from scratch, finetuning, and adoption as the feature extractor. The first scheme does not involve any external data, meaning there is no prior knowledge that can be leveraged. To remedy this issue, the finetuning scheme uses the networks pretrained on a large-scale dataset as the start point for further training (i.e., finetuning). The last scheme directly extracts the feature from the pretrained CNN without further finetuning, therefore lacking the flexibility of adapting to aerial images from different downstream tasks.

The existing literature \cite{asr_review} reveals that the finetuning strategy performs better than the other ones. We attribute it to the capacity of the used pretraining dataset, including the sample size and the number of categories. In the current RS field including the aerial scene recognition task, almost all finetuned models are pretrained on the ImageNet-1K dataset \cite{imagenet}, which is the most famous image dataset in the CV field. Its millions of real-world images from 1,000 different categories enable the models to learn a powerful representation. Usually, off-the-shelf deep models like VGG \cite{vgg} and ResNet \cite{resnet} are used as backbone networks for aerial scene recognition, since training a new network on the ImageNet from scratch is time-consuming and requires a large number of computational resources. To further improve the classification performance, some methods \cite{xu2021_dfagcn,sun_2019_GBNet} adopt the ImageNet pretrained models as the backbone network and employ multi-level features from it. In addition, many other components or strategies are specially designed for the aerial recognition task, such as the distillation \cite{ESDMBENet_2021_asr} and feature partitioning \cite{mgmlnet_asr_2021_featurepartioning}.

Although the aforementioned methods have achieved remarkable performance for aerial scene recognition, there are still some issues needed to be further investigated. Intuitively, when considering the characteristics of aerial images, there exists a large domain gap compared with the natural images in terms of view, color, texture, layout, object, etc. Previous methods attempt to narrow this gap by further finetuning the pretrained model on the RS image dataset. Nevertheless, the systematic bias introduced by ImageNet Pretraining (IMP) has a non-negligible side impact on the performance \cite{saumoco}. On the other hand, we notice that there are abundant aerial images captured by diverse multiple sensors with the progress of RS technology, which can be used for pretraining. As a representative example, MillionAID \cite{Long2021DiRS} is so far the largest aerial image dataset and has a million-level volume similar to ImageNet-1K, making the Remote Sensing Pretraining (RSP) become possible.

RSP enables training a deep model from scratch, implying that the candidate model is not necessary to be limited to the off-the-shelf CNN. In this paper, we also investigate the impact of RSP together with vision transformers, which have shown surprisingly good performance in the CV domain. Compared with convolutions in CNN which are skilled in modeling locality, the multi-head self-attention (MHSA) in transformers, e.g., Swin transformer \cite{swint}, can flexibly capture diverse global contexts. Recently, ViTAE \cite{xu2021vitae,vitae_v2} explores both convolutions and MHSA for modelling locality and long-range dependency simultaneously, achieving state-of-the-art (SOTA) performance on the ImageNet classification task and downstream vision tasks. In addition, it also extracts multi-scale features through a dilated convolution module and the stage-wise design, which have been shown effective in previous works, especially for aerial image interpretation \cite{wang_2021_asr_efpn}. Since both the CNN and aforementioned vision transformers can also produce intermediate features in different stages, which are useful for many downstream tasks, we also investigate their finetuning performance of them after RSP on semantic segmentation, object detection, and change detection. To achieve these goals, we conduct extensive experiments on nine popular datasets and have some findings. The RSP is an emerging research direction in aerial image understanding, which however is still underexplored, especially in the context of vision transformers. We hope this study could fill this gap and provide useful insights for future research.

The main contribution of this paper is three-fold.

\begin{itemize}
  
  \item[(1)] We investigate the impact of remote sensing pretraining by training on a large-scale remote sensing dataset using three types of backbone networks, including traditional CNN, competitive visual transformer models, and the advanced ViTAE transformers.
  \item[(2)] We further finetune the above models that are initialized with the remote sensing or ImageNet pretraining weights on four kinds of tasks including scene recognition, semantic segmentation, object detection, and change detection using a total of nine datasets, and compare them with other methods.
  \item[(3)] Experimental results show that typical vision transformer models can obtain competitive performance or perform better than CNN. Especially, the ViTAE achieves the best performance on almost all settings even if compared with the existing state-of-the-art methods. In addition, a series of findings of remote sensing pretraining will be presented, including the comparison with the traditional ImageNet pretraining and the performances on different downstream tasks. These findings provide useful insights for future research.
 \end{itemize}

 The remainder of this paper is organized as follows. Section II introduces the related works, including the aerial scene recognition methods, especially CNN and vision transformer related ones, and the existing works of RSP. Section III describes the implementations of RSP, as well as the employed large capacity MillionAID dataset and the adopted ViTAE network. The experiment results on the four tasks and the related comprehensive analyses are presented in section IV. Finally, Section V concludes this paper.

 \section{Related Work}

 \subsection{Aerial Scene Recognition}
 
 There are a large number of CNN-based approaches to aerial scene recognition. Many off-the-shelf CNN classification models that are pretrained on ImageNet, such as the VGG \cite{vgg}, ResNet \cite{resnet}, and DenseNet \cite{densenet}, have been used and further finetuned on aerial images. Nonetheless, the challenging aerial scene that possesses inter-class similarity and intra-class diversity can not be easily interpreted by only using features of the last layer, which are also considered as the ``global features'' compared with the features in previous layers, since it is also useful to highlight the important scene-relevant local regions for scene understanding. To address this issue, \cite{lse_2021_asr, xu2021_dfagcn} jointly exploits the multi-level CNN features, where the high-level features from deep layers usually have abundant semantic information, while the low-level features of the shallow layers tend to provide visual structures. For example, \cite{lse_2021_asr} conducts varied dilated convolutions on multiple features of VGG to obtain the more effective multiscale features. In addition, they optimize the category probabilities by preserving the local maximum and revising others with the two-dimensional Gaussian-like distribution in a window to strengthen the local regions. \cite{xu2021_dfagcn} separately applies graph convolutions on multilayer VGG features, where each pixel representation is regarded as a node, and the extracted graph features are concatenated with the last global visual features.

 Apart from feature fusion, the attention mechanisms have also been commonly used in aerial scene recognition since they can enhance the local features by simulating the human vision that directly assigns different weights to various areas of the current scene. The attention modules can be easily inserted into the CNN \cite{ma2020auto}. For example, \cite{mblanet_2021_asr} adopts channel attention and spatial attention modules in parallel like \cite{cbam} to form a complementary attention. \cite{lse_2021_asr} also employs spatial attention to further adjust the optimized category probabilities. Another point for aerial scene recognition is to model the relationships of different regions. For example, \cite{xu2021_dfagcn} captures the topological relations of different objects, where the adjacency matrices are carefully designed. Besides, some interesting topics such as the multiple instance learning \cite{midcnet_bi2020multiple_asr}, self-distillation \cite{ESDMBENet_2021_asr} and feature partition \cite{mgmlnet_asr_2021_featurepartioning} have also been explored in aerial scene recognition research.

 Networks before linear layers in scene recognition task can be used as feature encoders for many downstream tasks, where the most representative ones for aerial images are semantic segmentation, object detection, and change detection. While semantic segmentation and object detection are common tasks in CV, change detection is a specific task in RS. So far, a large number of related approaches in the aforementioned fields have been developed. Please refer to \cite{rs_review1,rs_review2,rs_review3,rs_review4} for details.

 \subsection{Vision Transformer}

 Transformer is first proposed in \cite{selfattention}, and has been widely used in the natural language processing (NLP) field \cite{nlp_review1,vision_transformer_review}. Besides NLP, the recently proposed vision transformers inspire a wave of researches \cite{vit,deit,pvt,swint,xu2021vitae,vitae_v2,segformer,zj_detection1,zj_detection2,zj_detection3,zj_detection5} in the CV field. The core component of the vision transformer is the MHSA, which is the extension of self-attention (SA). Compared with the convolution operation, the SA can capture long-range context and the relationships between any different positions. On the foundation of SA, the MHSA separately conducts the SAs in different projected subspaces, possessing more powerful representative abilities. ViT \cite{vit} is the pioneer vision transformer, where the input image is split into fixed-size patches to form tokens, which are then fed into the MHSA. However, the fixed receptive field restricts its applications on downstream tasks, and the global MHSA brings high computational complexities. To address the former issue, PVT \cite{pvt} adopts the classical pyramid structure, improving the model transferability with the generated hierarchical multiscale features. Swin \cite{swint} further substitutes the global MHSA to the shiftable window MHSA (WMHSA), it reduces the computational overhead significantly, achieving excellent performance on many CV tasks. Nonetheless, it still suffers from the common issues of vision transformers, such as being inefficient in modeling locality and scale invariance, which are exactly the advantages of CNN. Thus, besides Swin, we also employ another advanced vision transformer named ViTAE \cite{xu2021vitae,vitae_v2}, where the intrinsic biases of CNN are introduced into the transformer. It also adopts a pyramid structure to generate hierarchical features and local window attention, achieving better performance on many CV tasks while having a reasonable computational overhead and memory footprint. Using vision transformers as the backbone for aerial scene recognition is still under-explored, while the existing method merely takes ViT as a branch parallel to CNN \cite{cnn_transformer_2022_asr}, implying an urgent need to further explore the applications of vision transformers on aerial scene tasks.

 \subsection{Remote Sensing Pretraining}

 \begin{figure*}[h]
  \centering
  \includegraphics[width=1\linewidth]{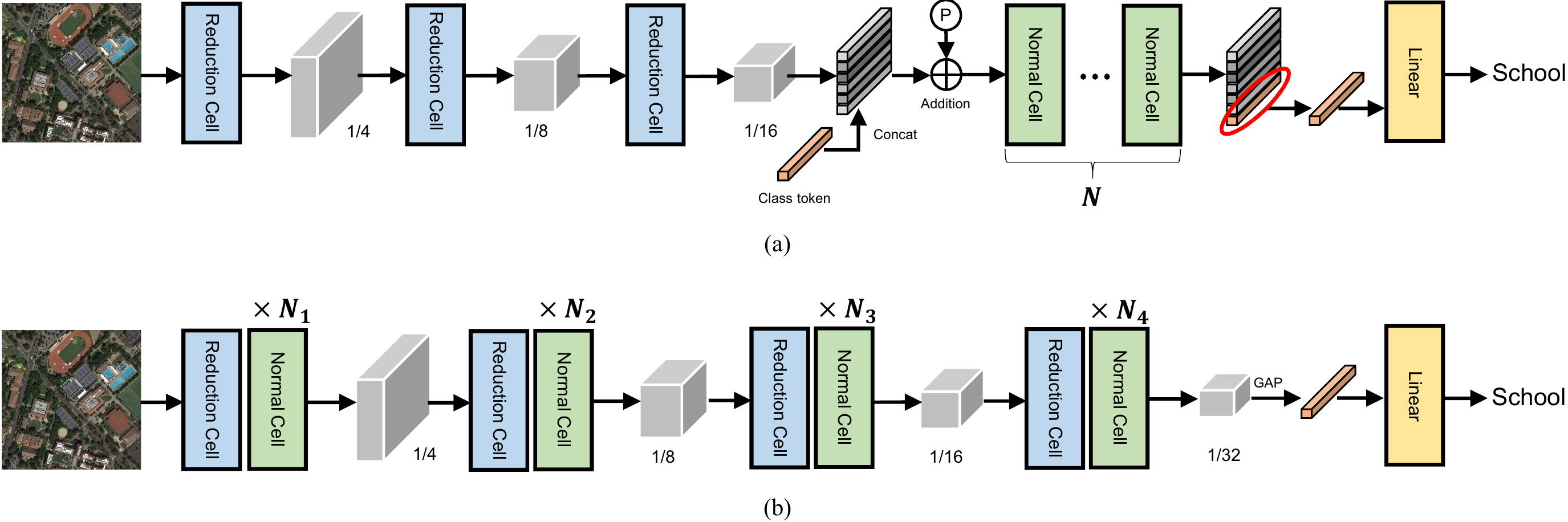}\\
  \caption{The diagram of the adopted ViTAE models. (a) Original ViTAE \cite{xu2021vitae}. (b) ViTAEv2 \cite{vitae_v2}.}
  \label{vitae}
\end{figure*}

 Pretraining using RS dataset for aerial scene recognition is a very intuitive idea. However, to our best knowledge, there are few explorations in this direction since the insufficiency of large-scale RS datasets like ImageNet. Nevertheless, researchers have attempted to obtain the RS representations from other resources. For example, GeoKR \cite{geokr} leverages the global land cover product as the labels, and they use the mean-teacher framework to alleviate the influences of imaging time and resolution differences between RS images and geographical ones. However, forcing alignment of different datasets inevitably brings errors due to the intrinsic different data distributions. The scarcity of large capacity RS dataset is mainly in the aspect of category labels instead of images. In this case, it is promising to develop self-supervised pretraining methods \cite{simclr, mocov2, regioncl, mae} and some related methods have been developed in the RS area \cite{seco, saumoco, geography_aware, hyperspectral_color}. For instance, SeCo \cite{seco} leverages the seasonal changes to enforce consistency between positive samples, which are the unique characteristics of aerial scenes, while \cite{geography_aware} simultaneously fuses the temporal information and geographical location into the MoCo-V2 \cite{mocov2} framework. Moreover, the channel properties \cite{hyperspectral_color} and spatial variance \cite{saumoco} are also explored in some approaches. In this study, since the adopted MillionAID dataset has the ground truth labels annotated by experts and does not contain any temporal information, we directly conduct the supervised pretraining like the conventional IMP

 \section{Remote Sensing Pretraining}

 In this section, we first provide a brief introduction of the adopted large-scale RS dataset---MillionAID. Then, we describe the details of the utilized ViTAE transformer. The whole RSP procedure will be presented finally.
 
 \subsection{MillionAID}
 
 To our best knowledge, the MillionAID is by far the largest dataset in the RS area. It contains 100,0848 non-overlapping scenes, exceeding the competitive counterpart fMoW \cite{fmow} and BigEarthNet \cite{bigearthnet}, which separately includes 132,716 and 590,326 scenes. Note that the fMoW contains 1,047,691 images since they provide multiple temporal views for each scene. In addition, it should be noted that the fMoW and BigEarthNet are multispectral datasets, while the MillionAID is an RGB dataset, which is more suitable for existing deep vision models. The categories of MillionAID consist of a hierarchical tree that has 51 leaves, which locate on 28 parent nodes on the second level, while the 28 groups are belonging to 8 base classes: agriculture land, commercial land, industrial land, public service land, residential land, transportation land, unutilized land, and water area, and each leaf category has about 2,000$\sim$45,000 images. This dataset is obtained from the Google Earth that is made up of diverse sensors including but not limited to SPOT, IKONOS, WorldView, and Landsat series, resulting in different resolutions. The maximum resolution can reach 0.5m, while the smallest is 153m. The image size ranges from 110 $\times$ 110 to 31,672 $\times$ 31,672.

 \subsection{ViTAE}

The original ViTAE \cite{xu2021vitae} follows the deep-narrow design of T2T-ViT \cite{t2tvit}, which found that simply decreasing channel dimensions and increasing layer depth can improve the feature richness of ViT, boosting the performance while reducing the model size and computational cost. Thus, the original ViTAE firstly downsamples the input image to 1/16 size by three reduction cells. Similar to ViT, a class token is concatenated with the output of the third reduction cell before adding an element-wise sinusoid position encoding. Then, multiple normal cells are stacked, and the feature size is always kept till the end. The class token feature of the last normal cell is used for classification through a linear layer.

Although the original ViTAE performs well on ImageNet classification, it is unsuitable for transferring to other tasks like segmentation, detection, pose estimation, and so on, since it cannot generate abundant intermediate features at different scales. Thus, the authors propose the ViTAEv2 variant \cite{vitae_v2}, which adopts the classic stage-wise design of popular backbone networks such as ResNet and Swin. Figure \ref{vitae} shows the comparison between the original ViTAE and ViTAEv2. In ViTAEv2, the network is split into multiple stages, usually, the number is 4. In each stage, the first cell is a reduction cell for downsampling, which is followed by the stacked normal cells. An average pooling layer is used after the last normal cell to replace the class tokens. When finetuning on downstream tasks, this pooling layer is removed and the remained network is connected with corresponding task decoders.

\begin{figure}[t]
  \centering
  \includegraphics[width=1\linewidth]{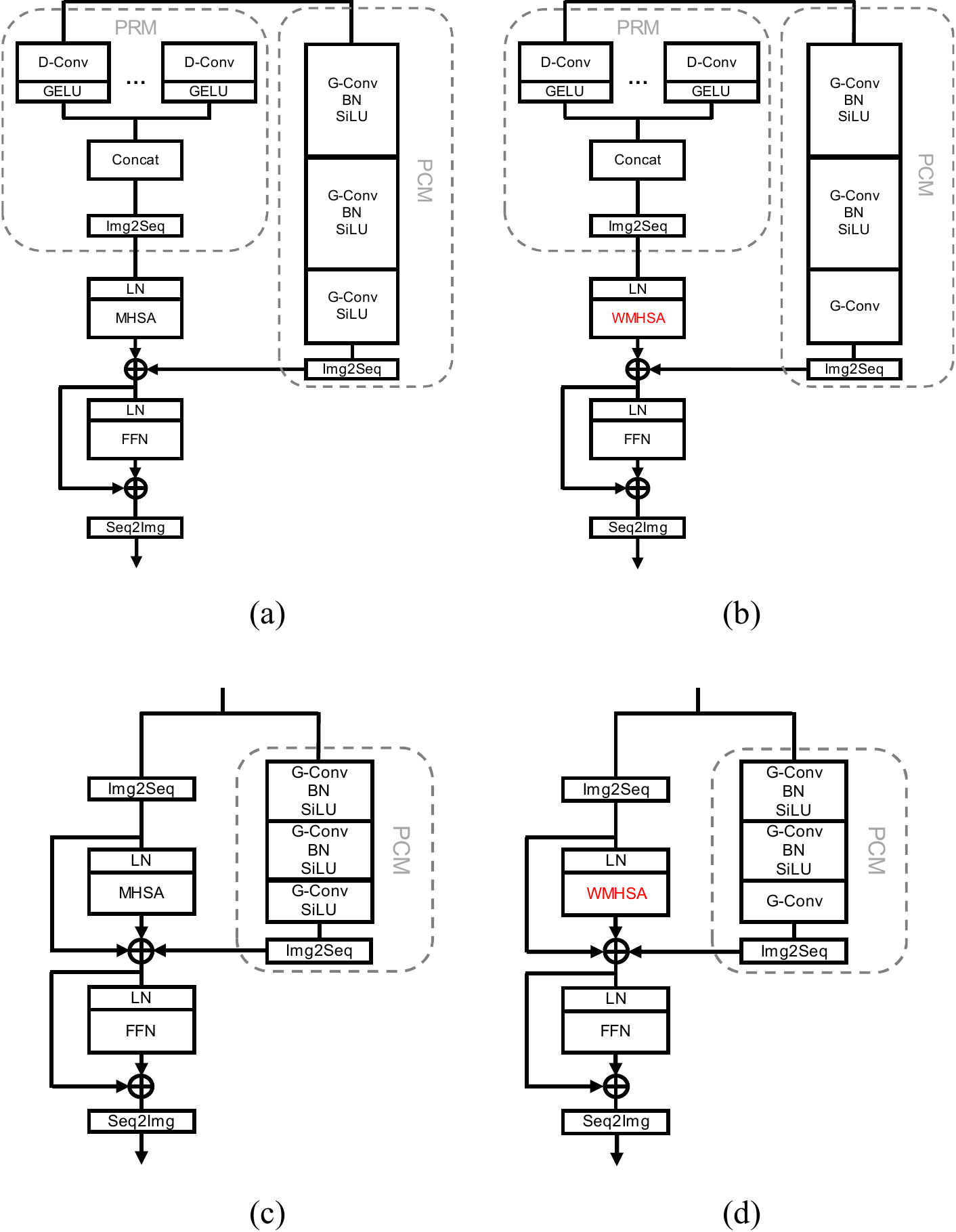}\\
  \caption{The structures of different cells in ViTAE models. (a) and (c) are the reduction cell and normal cell in the original ViTAE, while (b) and (d) are the corresponded variants in the ViTAEv2.
  }
  \label{cell}
\end{figure}

In this paper, we employ the ViTAEv2 model for RSP. Concretely, inspired by Swin \cite{swint}, some MHSAs in ViTAE are replaced by the WMHSA to reduce computational cost. Specifically, considering the feature size becomes smaller in later stages, it is unnecessary to partition the feature for WMHSA. Thus, only the MHSAs in the first two stages are substituted by WMHSA. It should be noticed that the adopted WMHSA does not need to be shifted as the original implementation, since the WMHSA is conducted on the merged multiscale feature from the pyramid reduction module (PRM), where different regions have communicated with each other through the overlapped receptive fields of the sliding dilated convolutions. Besides, it is also not necessary to use relative positional encoding since the convolutions already encode the positional information. Additionally, the SiLU \cite{silu} in the last convolutional layer of the parallel convolutional module (PCM) is also removed to reduce nonlinearity. The structures and comparisons of different cells in the original ViTAE and ViTAEv2 have been shown in Figure \ref{cell}. For reduction cell, normal cell, PRM and PCM, readers can refer to \cite{xu2021vitae} and \cite{vitae_v2} for more details.

\begin{table}[t]
  \scriptsize
  \caption{The hyperparameter settings of different ``small'' version ViTAE models. ``P'' denotes Performer attention \cite{t2tvit} while ``L'' means the reduction cell has no PCM and attention. ``F'' denotes the original MHSA while ``W'' denotes the WMHSA.}
  \newcommand{\tabincell}[2]{\begin{tabular}{@{}#1@{}}#2\end{tabular}}
  \centering
  \resizebox{\linewidth}{!}{
  \begin{tabular}{l|l|l|l}
  \hline
  \multicolumn{2}{l|}{Network} & ViTAE-S \cite{xu2021vitae} & ViTAEv2-S \cite{vitae_v2} \\
  \hline
  \multicolumn{2}{l|}{Stage} & 3 & 4\\
  \multicolumn{2}{l|}{Downsampling Ratio} & [4, 2, 2] & [4, 2, 2, 2] \\
  \multicolumn{2}{l|}{Embedding Dim} & [64, 64, 192] & [64, 64, 128, 256] \\
  \multicolumn{2}{l|}{Stage Dim} & [96, 192, 384] & [64, 128, 256, 512] \\
  \hline
  \multirow{3}{*}{RC} &
  Head & [1, 1, 1] & [1, 1, 2, 4] \\
  &Group & [1, 1, 1] & [1, 16, 32, 64] \\
  & Type & [P, P, L] & [W, W, F, F] \\
  \hline
  \multirow{4}{*}{NC} &
  Head & [1, 1, 6] & [1, 2, 4, 8] \\
  & Group & [1, 1, 96] & [1, 32, 64, 128] \\
  & Type & [F, F, F] & [W, W, F, F] \\
  & Depth & [0, 0, 14] & [2, 2, 8, 2]  \\
  \hline
\end{tabular}
  }
  \label{three_networks}
\end{table}

\begin{table*}[t]
  \scriptsize
  \caption{Results of different models on the mini-evaluation set. They are trained on the mini-training set from MillionAID.}
  \newcommand{\tabincell}[2]{\begin{tabular}{@{}#1@{}}#2\end{tabular}}
  \centering
  \begin{tabular}{l|c|c|c|c}
  \hline
  Model & Acc@1 & Acc@5 & Param(M) & Training Time (hh:mm:ss) \\ 
  \hline
  ResNet-50 \cite{resnet} & 80.8 & 96.1 & 23.6 & 16:42:58 \\
  DeiT-S \cite{deit} & 72.9 &  92.9 & 21.7 & 16:57:39 \\
  ViT-B \cite{vit} & 78.4 & 92.1 & 114.4 & 17:41:03 \\
  PVT-S \cite{pvt} & 80.0 & 94.1 & 23.9 & 18:06:42 \\
  Swin-T \cite{swint} & 84.7 & 93.7 & 27.6 & 17:49:06 \\
  ViTAE-S \cite{xu2021vitae} & 85.9 & 96.9 & 22.8 & 24:00:35 \\
  ViTAEv2-S \cite{vitae_v2} & 88.2 & 96.1 & 18.8 & 23:57:30 \\ 
  \hline
\end{tabular}
  \label{model_select}
\end{table*}

In our implementation, we mainly evaluate the ``small'' version of the original ViTAE, named ViTAE-S. In addition, we also adopt the ViTAEv2-S model due to its excellent representation ability and transferability to downstream tasks. Table \ref{three_networks} lists the details of ViTAE-S and ViTAEv2-S. Here, the length of the corresponded list equals the number of stages. ``Embedding Dim'' means the encoding dimension in PRM, while ``Stage Dim'' is the channel number of the feature through the corresponding stage, which is useful for aligning the related downstream task decoders. The ``RC'' and ``NC'' separately represent the reduction cell and normal cell, where ``Head'' is the head number in MHSA or WMHSA, ``Group'' represents the number of group convolutions in PCM, and ``Type'' is the specific attention types. The ViTAE-S adopts the performer of T2T-ViT \cite{t2tvit} in the first two reduction cells. ``L'' means this reduction cell does not use PCM and attention. ``F'' and ``W'' separately denote the MHSA and WMHSA in ViT and Swin. At last, the ``Depth'' is the number of stacked normal cells, which is also the $N_i$ in Figure \ref{vitae}.

\subsection{Implementation}
\subsubsection{Determine the Pretraining Network}
\label{subsubsec:network}

We first determine the type of deep models to be used for RSP. To this end, from the official training set, we construct a mini-training set and a mini-evaluation set, which have 9,775 and 225 images, respectively. Note the latter set is formed by randomly selecting 5 images from each category to balance the classes. For CNN, the classic ResNet-50 \cite{resnet} is employed. Since this research mainly explores the performance of vision transformer models with RSP, a series of typical vision transformer based networks including DeiT-S \cite{deit}, PVT-S \cite{pvt}, and Swin-T \cite{swint}, are also evaluated. The selection of a specific version is to guarantee these models have a similar amount of parameters compared with the ViTAE-S model. In addition, we also include ViT-B \cite{vit} for reference since ViT is the most basic model of vision transformers. 

All models are trained with 300 epochs and batch size 16. We adopt the AdamW optimizer where the momentum is set to 0.9 and the weight decay is set to 0.05. The initial learning rate is 1e-3, which is adjusted by the cosine scheduling policy: $current\_lr = min\_lr+\cfrac{1}{2}(initial\_lr-min\_lr)(1+\cos(\frac{iter}{max\_iter}\pi))$, where $min\_lr$ is 5e-6. In addition, we set the warming up epochs to 5, where the learning rate is set to 5e-7. Following the typical IMP, the input image is resized to 224 $\times$ 224 by randomly cropping during training, while during testing, the image at the same size is obtained through ``center crop''. In addition, a series of data argumentations including AutoAugment \cite{autoaugment}, Random Erasing \cite{random_erase}, Mixup \cite{mixup}, CutMix \cite{cutmix}, and color jitter are applied to improve the training performance. The top-1 accuracy and top-5 accuracy are used as the evaluation metrics. In addition, all models are implemented on a single NVIDIA Tesla V100 GPU, and the results are shown in Table \ref{model_select}.

It can be seen that despite the ViT-B has the most parameters, it performs not better than the classic ResNet-50. The DeiT-S performs the worst since we do not adopt the teacher model. Because our task is pretraining using RS images, obtaining the corresponding teacher model is our target instead of the prerequisite. By introducing the design paradigm of the feature pyramid, PVT-S improves the accuracy compared with ViT-B. On this foundation, the original ViTAE-S further considers the locality and scale-invariance modeling, which are the inductive biases of conventional CNN models. However, it cost much training time since the token number in the early RCs is large, requiring more computations. The Swin-T addresses this issue by restricting the MHSA in fixed windows and adopts the image shifting to implicitly promote the communications between windows. By taking the advantage of WMHSA, the ViTAEv2-S achieves the best performance and it exceeds the second place by 2.3\% top-1 accuracy.

The procedure of model determination is shown as follows. For the ViTAE models, we choose the strongest one to expect good performance in downstream tasks such as the aerial scene recognition when adopting RSP, i.e., the ViTAEv2-S. For comparison, the ResNet-50 is selected as the representative network in conventional CNN, and the RS pretrained ResNet-50 can also provide a group of new CNN related baselines on a series of aerial datasets. The DeiT-S and ViT-B are eliminated because of the low accuracy and a large number of parameters, and they are difficult to be transferred into downstream tasks because of the design of stacking transformers. The Swin can be regarded as building on the foundation of PVT by substituting the global MHSA with the shiftable WMHSA. Since the top-1 accuracy of Swin is larger than PVT, and Swin-T requires less training time, we also choose Swin-T in the subsequent experiments.

\subsubsection{Obtain the Suitable Weights}

After determining the model candidates, we conduct the RSP to obtain the pretrained weights. Concretely, maintaining the category balance, we randomly choose 1,000 images in each category of the MillionAID dataset to form the validation set that has 51,000 images, achieving a similar volume compared with the ImageNet validation set, which contains 50,000 images. The remaining 949,848 images are used for training. Although the number of images and categories for RSP are less than those of the ImageNet training set, it still can perform to be competitive or even achieves SOTA results on aerial scene tasks, whose details will be presented later.

\begin{table}[t]
  \scriptsize
  \caption{Results of ViTAEv2-S with different settings of training epoch on the MillionAID validation set.}
  \newcommand{\tabincell}[2]{\begin{tabular}{@{}#1@{}}#2\end{tabular}}
  \centering
  \begin{tabular}{l|c|c}
  \hline
  Epoch & Acc@1 & Acc@5\\
  \hline
  5 & 94.53 & 99.41 \\
  10 & 96.45 & 99.64 \\
  15 & 97.38 & 99.74 \\
  20 & 98.00 & 99.81 \\
  40 & 98.64 & 99.86 \\
  60 & 98.87 & 99.83 \\
  80 & 98.90 & 99.85 \\
  100 & 98.97 & 99.88\\
  \hline
\end{tabular}
  \label{vitae_rsp}
\end{table}

\begin{table}[t]
  \scriptsize
  \caption{Results of the candidate models for the subsequent finetuning experiments on the MillionAID validation set.}
  \newcommand{\tabincell}[2]{\begin{tabular}{@{}#1@{}}#2\end{tabular}}
  \centering
  \begin{tabular}{l|c|c}
  \hline
  Epoch & Acc@1 & Acc@5 \\
  \hline
  \bfseries \textit{ResNet-50} & \multicolumn{2}{c}{}\\
  \hline
  40 & 97.99 & 99.81  \\
  120 & 98.76 & 99.83 \\
  300 & 98.99 & 99.82 \\
  \hline
  \bfseries \textit{Swin-T} & \multicolumn{2}{c}{}\\
  \hline
  40 & 97.80 & 99.84  \\
  120 & 98.63 & 99.89 \\
  300 & 98.59 & 99.88  \\
  \hline
  \bfseries \textit{ViTAEv2-S} & \multicolumn{2}{c}{}\\
  \hline
  40 & 98.64 & 99.86 \\
  100 & 98.97 & 99.88 \\
  \hline
\end{tabular}
  \label{final_models}
\end{table}

To obtain suitable pretraining weights, we separately train the ViTAEv2-S model under the configuration of different epochs. The basic learning rate is 5e-4, and the batch size is set to 384. The remained settings are the same as the previous experiment. All experiments are conducted with 4 V100 GPUs, and the results are shown in Table \ref{vitae_rsp}. According to the results, it can be observed that the model starts saturation after about 40 epochs, since it only improves 0.64\% top-1 accuracy compared with training 20 epochs, while the next 20 epochs only bring a gain of 0.23\%. Thus, the network weights trained with 40 epochs are firstly chosen as the RSP parameters of ViTAEv2-S to be applied to the subsequent tasks. Intuitively, the model achieving good performance on the large-scale pretraining dataset will also perform well on the downstream tasks. Therefore, we also use the network weights trained with 100 epochs in the downstream tasks. These models are separately denoted with the suffix ``E40'' and ``E100''.

For ResNet-50 and Swin-T, we follow \cite{swint} to configure the training settings, where the networks are trained for 300 epochs. In the experiments, we observe that the top-1 accuracy of Swin-T-E120 on the validation set is roughly equivalent to ViTAEv2-S-E40. Thus, the training weights of Swin-T-E120 are selected. Similarly, we also choose the final network weights Swin-T-E300 as a comparison with ViTAEv2-S-E100. To make the experiments fair, the weights of ResNet-50 and Swin-T that are trained with 40 epochs are also considered, since they are trained using the same number of epochs with the ViTAEv2-S-E40.

The final pretraining models are listed in Table \ref{final_models}. It can be seen that the validation accuracies are almost increasing with the increase of training epochs. However, the performance of Swin-T-E300 is not as well as Swin-T-E120. Nonetheless, we still keep it since it may have stronger generalization by seeing more diverse samples.

\section{Finetuning on Downstream Tasks}

In this section, the pretrained models are further finetuned on a series of downstream tasks, including recognition, semantic segmentation, object detection in aerial scenes as well as change detection. It should be clarified that models for scene recognition in this section are trained and evaluated on commonly used aerial scene datasets rather than the MillionAID engaging for RSP.

\subsection{Aerial Scene Recognition}

We first introduce the used scene recognition datasets and the implementation details, then present the experimental results and analyses.

\subsubsection{Dataset}

The three most popular scene recognition datasets including the UC Merced Land Use (UCM) dataset \cite{ucm}, the Aerial Image Dataset (AID) \cite{aid}, and the benchmark for RS Image Scene Classification that is created by Northwestern Polytechnical University (NWPU-RESISC) \cite{asr_review}, are used to comprehensively evaluate the impact of RSP and the representation ability of the above adopted backbones.

\begin{itemize}
  \item UCM: This is the most important dataset for scene recognition. It contains 2,100 images whose sizes are all 256 $\times$ 256 and have a pixel resolution of 0.3m. The 2,100 images equally belong to 21 categories. Thus, each category has 100 images. All samples are manually extracted from the large images in the USGS National Map Urban Area Imagery Database collected from various urban areas around the country.
  \item AID: This is a challenging dataset, which is generated by collecting the images from multi-source sensors on GE. It has high intra-class diversities since the images are carefully chosen from different countries. And they are extracted at different times and seasons under different imaging conditions. It has 10,000 images at the size of 600 $\times$ 600, belonging to 30 categories.
  \item NWPU-RESISC: This dataset is characterized by a great number of samples. It contains 31,500 images and 45 categories in total, where each category has 700 samples. Each image has 256 $\times$ 256 pixels. The spatial resolutions are varied from 0.2m to 30m. Some special landforms, such as islands, lakes, regular mountains, and snow mountains, maybe in lower resolutions.
\end{itemize}

\subsubsection{Implementation Detail and Experimental Setting}

\begin{table*}[t]
  \scriptsize
  \caption{Results of the selected models and SOTA methods on the three scene recognition datasets under different settings. The bold fonts in the last three groups mean the best results, while ``\textbf{*}'' denotes the best among all models.}
  \newcommand{\tabincell}[2]{\begin{tabular}{@{}#1@{}}#2\end{tabular}}
  \centering
  \resizebox{\linewidth}{!}{
  \begin{tabular}{l|c|c|c|c|c|c}
  \hline
  Model & Publication & UCM (8:2) & AID (2:8) & AID (5:5) & NWPU-RESISC (1:9) & NWPU-RESISC (2:8) \\
  \hline
  CBAM \cite{cbam}  & ECCV2018 & 99.04 $\pm$ 0.23 & 94.66 $\pm$ 0.39 & 96.90 $\pm$ 0.04 & 92.10 $\pm$ 0.04 & 94.26 $\pm$ 0.12 \\
  EAM (IMP-ResNet-50) \cite{asr_2021_grsl_eam} & GRSL2021 & 98.98 $\pm$ 0.37 & 93.64 $\pm$ 0.25 & 96.62 $\pm$ 0.13 & 90.87 $\pm$ 0.15 & 93.51 $\pm$ 0.12 \\
  $\text{F}^2$BRBM (IMP-ResNet-50) \cite{asr_2021_jstars_f2brbm} &JSTARS2021& 99.58 $\pm$ 0.23 & 96.05 $\pm$ 0.31 & 96.97 $\pm$ 0.22 & 92.74 $\pm$ 0.23 & 94.87 $\pm$ 0.15 \\
  MBLANet (IMP-ResNet-50) \cite{mblanet_2021_asr} & TIP2021 & 99.64 $\pm$ 0.12 & 95.60 $\pm$ 0.17 & 97.14 $\pm$ 0.13 & 92.32 $\pm$ 0.15 & 94.66 $\pm$ 0.11 \\
  GRMANet (IMP-ResNet-50) \cite{asr_2022_tgrs_grmanet} & TGRS2021 & 99.19 $\pm$ 0.10 & 95.43 $\pm$ 0.32 & 97.39 $\pm$ 0.24 & 93.19 $\pm$ 0.42 & 94.72 $\pm$ 0.25 \\
  IDCCP (IMP-ResNet-50) \cite{asr_2021_tgrs_idccp}& TGRS2021 & 99.05 $\pm$ 0.20 & 94.80 $\pm$ 0.18 & 96.95 $\pm$ 0.13 & 91.55 $\pm$ 0.16 & 93.76 $\pm$ 0.12 \\
  ESD-MBENet-v1 (IMP-ResNet-50) \cite{ESDMBENet_2021_asr}& TGRS2021 & 99.81 $\pm$ 0.10 & 96.00 $\pm$ 0.15 & 98.54 $\pm$ 0.17 & 92.50 $\pm$ 0.22 & 95.58 $\pm$ 0.08 \\ 
  ESD-MBENet-v2 (IMP-ResNet-50) \cite{ESDMBENet_2021_asr}& TGRS2021 & 99.86 $\pm$ 0.12 & 95.81 $\pm$ 0.24 & 98.66 $\pm$ 0.20 & 93.03 $\pm$ 0.11 & 95.24 $\pm$ 0.23 \\
  \hline
  ARCNet (IMP-VGG-16) \cite{asr_2019_tgrs_arcnet} & TGRS2019 & 99.12 $\pm$ 0.40 & 88.75 $\pm$ 0.40 & 93.10 $\pm$ 0.55 & --- & --- \\
  SCCov (IMP-VGG-16) \cite{asr_2020_tnnls_sccov} & TNNLS2019 & 99.05 $\pm$ 0.25 & 93.12 $\pm$ 0.25 & 96.10 $\pm$ 0.16 &89.30 $\pm$ 0.35 & 92.10 $\pm$ 0.25 \\
  KFBNet (IMP-DenseNet-121) \cite{asr_2020_tgrs_kfbnet}& TGRS2020 & 99.88 $\pm$ 0.12 & 95.50 $\pm$ 0.27 & 97.40 $\pm$ 0.10 & 93.08 $\pm$ 0.14 & 95.11 $\pm$ 0.10 \\
  GBNet (IMP-VGG-16) \cite{sun_2019_GBNet} & TGRS2020 & 98.57 $\pm$ 0.48 & 92.20 $\pm$ 0.23 & 95.48 $\pm$ 0.12 & --- & --- \\
  MG-CAP (IMP-VGG-16) \cite{asr_2020_tip_mgcap} & TIP2020 & 99.00 $\pm$ 0.10 & 93.34 $\pm$ 0.18 & 96.12 $\pm$ 0.12 & 90.83 $\pm$ 0.12 & 92.95 $\pm$ 0.13 \\
  EAM (IMP-ResNet-101) \cite{asr_2021_grsl_eam} & GRSL2021 & 99.21 $\pm$ 0.26 & 94.26 $\pm$ 0.11 & 97.06 $\pm$ 0.19 & 91.91 $\pm$ 0.22 & 94.29 $\pm$ 0.09 \\
  IMP-ViT-B \cite{vit} & ICLR2021 & 99.28 $\pm$ 0.23 & 93.81 $\pm$ 0.21 & 96.08 $\pm$ 0.14 & 90.96 $\pm$ 0.08 & 93.96 $\pm$ 0.17 \\
  MSANet (IMP-ResNet-101) \cite{asr_2021_jstars_msanet}  & JSTARS2021 & 98.96 $\pm$ 0.21 & 93.53 $\pm$ 0.21 & 96.01 $\pm$ 0.43 & 90.38 $\pm$ 0.17 & 93.52 $\pm$ 0.21 \\
  CTNet (IMP-MobileNet-V2+IMP-ViT-B) \cite{cnn_transformer_2022_asr} & GRSL2021 & --- & 96.25 $\pm$ 0.10 & 97.70 $\pm$ 0.11 & 93.90 $\pm$ 0.14 & 95.40 $\pm$ 0.15 \\
  LSENet (IMP-VGG-16) \cite{lse_2021_asr} & TIP2021 & 99.78 $\pm$ 0.18 & 94.41 $\pm$ 0.16& 96.36 $\pm$ 0.19 & 92.23 $\pm$ 0.14 & 93.34 $\pm$ 0.15 \\ 
  DFAGCN (IMP-VGG-16) \cite{xu2021_dfagcn} & TNNLS2021 & 98.48 $\pm$ 0.42 & --- & 94.88 $\pm$ 0.22 & --- & 89.29 $\pm$ 0.28 \\
  MGML-FENet (IMP-DenseNet-121) \cite{mgmlnet_asr_2021_featurepartioning}& TNNLS2021 & 99.86 $\pm$ 0.12 & 96.45 $\pm$ 0.18 & 98.60 $\pm$ 0.04 & 92.91 $\pm$ 0.22 & 95.39 $\pm$ 0.08 \\
  ESD-MBENet-v1 (IMP-DenseNet-121) \cite{ESDMBENet_2021_asr}& TGRS2021 & 99.86 $\pm$ 0.12 & 96.20 $\pm$ 0.15 & 98.85 $\pm$ 0.13\textbf{*} & 93.24 $\pm$ 0.15 & 95.50 $\pm$ 0.09 \\ 
  ESD-MBENet-v2 (IMP-DenseNet-121) \cite{ESDMBENet_2021_asr}& TGRS2021 & 99.81 $\pm$ 0.10 & 96.39 $\pm$ 0.21 & 98.40 $\pm$ 0.23 & 93.05 $\pm$ 0.18 & 95.36 $\pm$ 0.14 \\
  \hline
  IMP-ResNet-50 \cite{resnet} & CVPR2016 & 98.81 $\pm$ 0.23 & 94.67 $\pm$ 0.15 & 95.74 $\pm$ 0.10 & 90.09 $\pm$ 0.13 & 94.10 $\pm$ 0.15 \\
  SeCo-ResNet-50 \cite{seco} & ICCV2021 & 97.86 $\pm$ 0.23 & 93.47 $\pm$ 0.08 & 95.99 $\pm$ 0.13 & 89.64 $\pm$ 0.17 & 92.91 $\pm$ 0.13 \\
  RSP-ResNet-50-E40 &  Ours  & 99.43 $\pm$ 0.24 & 95.88 $\pm$ 0.07 & 97.29 $\pm$ 0.07 & 92.86 $\pm$ 0.09 & 94.40 $\pm$ 0.05 \\
  RSP-ResNet-50-E120 & Ours & \bfseries 99.52 $\pm$ 0.15 &  96.60 $\pm$ 0.04 & 97.78 $\pm$ 0.08 & 93.76 $\pm$ 0.03 & 94.97 $\pm$ 0.07\\
  RSP-ResNet-50-E300 & Ours & 99.48 $\pm$ 0.10 & \bfseries 96.81 $\pm$ 0.03 & \bfseries 97.89 $\pm$ 0.08 & \bfseries 93.93 $\pm$ 0.10 & \bfseries 95.02 $\pm$ 0.06\\
  \hline
  IMP-Swin-T \cite{swint} & ICCV2021 & \bfseries 99.62 $\pm$ 0.19 & 96.55 $\pm$ 0.03 & 98.10 $\pm$ 0.06 & 92.73 $\pm$ 0.09 & \bfseries 94.70 $\pm$ 0.10 \\
  RSP-Swin-T-E40 & Ours & 99.24 $\pm$ 0.18 & 95.95 $\pm$ 0.06 & 97.52 $\pm$ 0.04  & 91.22 $\pm$ 0.18 & 93.30 $\pm$ 0.08 \\
  RSP-Swin-T-E120 & Ours & 99.52 $\pm$ 0.00 & 96.73 $\pm$ 0.07 & 98.20 $\pm$ 0.02 & 92.02 $\pm$ 0.14 & 93.84 $\pm$ 0.07 \\
  RSP-Swin-T-E300 & Ours & 99.52 $\pm$ 0.00 & \bfseries 96.83 $\pm$ 0.08 &\bfseries 98.30 $\pm$ 0.04 & \bfseries 93.02 $\pm$ 0.12 & 94.51 $\pm$ 0.05 \\
  \hline
  IMP-ViTAEv2-S \cite{vitae_v2} & arXiv2022 & 99.71 $\pm$ 0.10 & 96.61 $\pm$ 0.07 & 98.08 $\pm$ 0.03 & 93.90 $\pm$ 0.07 & 95.29 $\pm$ 0.12 \\ 
  RSP-ViTAEv2-S-E40 & Ours & 99.71 $\pm$ 0.10 & 96.72 $\pm$ 0.06 & 97.92 $\pm$ 0.06 & 94.12 $\pm$ 0.07& 95.35 $\pm$ 0.03\\
  RSP-ViTAEv2-S-E100 & Ours & \bfseries 99.90 $\pm$ 0.13\textbf{*} & \bfseries 96.91 $\pm$ 0.06\textbf{*} & \bfseries 98.22 $\pm$ 0.09 &\bfseries 94.41 $\pm$ 0.11\textbf{*} & \bfseries 95.60 $\pm$ 0.06\textbf{*} \\
  \hline
\end{tabular}
  }
  \label{asr}
\end{table*}

The training settings are the same as previous experiments. The training epoch and batch size are set to 200 and 64, respectively. These experiments are conducted on a single V100 GPU. Following \cite{mblanet_2021_asr}, five settings of these three datasets are adopted to comprehensively evaluate the RS pretrained models and make the experiments become convincible, including UCM (8:2), AID (2:8), AID (5:5), NWPU-RESISC (1:9), and NWPU-RESISC (2:8). Note the $m:n$ means $10m\%$ samples are used for training, while the others form the testing set. Similar to the previous section, the images in each category are proportionally divided into two groups that are separately used for training and evaluation, respectively. Besides the above three backbones we selected, the ImageNet pretrained ResNet-50 and the ResNet-50 pretrained by SeCo \cite{seco} -- an RS self-supervised method considering seasonal variation, are also adopted for a fair comparison. When implementing finetuning on each scene recognition task, only the neuron number of the last linear layer is changed to match the categories of the target dataset. The overall accuracy (OA), which is the most commonly used criterion in the aerial scene recognition community by counting the proportion of the correct classified images relative to all images in the testing set, is utilized in the experiments. The models are repeatedly trained and evaluated five times at each setting, and the average value $\mu$ and standard deviation $\sigma$ of the results in different trials are recorded as $\mu \pm \sigma$.

\subsubsection{Experimental Results}
\textbf{Quantitative Results and Analyses:} Table \ref{asr} presents the results of the above selected backbones pretrained using different methods and other SOTA methods. Since this research only focuses on the pretraining of deep networks, especially the vision transformers. We only lists the DL based aerial scene recognition methods. For convenience, the ``IMP'' and ``RSP'' are used to represent ``ImageNet Pretraining'' and ``Remote Sensing Pretraining'', respectively. It can be seen that the methods are split into five groups. The first group is the methods that adopt ResNet-50 as the backbone network, where the ResNet-50 is initialized by the ImageNet pretrained weights. This group can be used to compare with the third group. The second group includes the recent existing advanced methods whose backbone is other popular networks except for ResNet-50, such as the ImageNet pretrained VGG-16, ResNet-101, DenseNet-121, and so on. Then, the ResNet-50, Swin-T, and ViTAEv2-S networks, whose pretrained weights are obtained by IMP, RSP, or SeCo, form the last three groups, respectively. In addition, it should be noted that besides the network types, the weights pretrained for different epochs are also considered. The bold fonts in the last three groups mean the best results in each group, while ``\textbf{*}'' denotes the best among all models (same meanings in other tasks).

On the foundation of ImageNet pretrained ResNet-50, many methods are developed, which have been shown in the first group. Among these methods, many flexible and advanced modules have been explored. For example, the attention mechanisms (CBAM \cite{cbam}, EAM \cite{asr_2021_grsl_eam}, MBLANet \cite{mblanet_2021_asr}), where specific channels or spatial positions of the features are highlighted, and multiscale features ($\text{F}^2$BRBM \cite{asr_2021_jstars_f2brbm} and GRMANet \cite{asr_2022_tgrs_grmanet}), where the intermediate features are also employed. In addition, the self-distillation technology combined with specially designed loss functions (ESD-MBENet \cite{ESDMBENet_2021_asr}) and the multibranch siamese networks (IDCCP \cite{asr_2021_tgrs_idccp}) have also been applied. While in the second group, the more diverse frameworks with various backbones are presented. Besides the traditional CNN, the recent ViT has also been applied in some works. Compared with the IMP-ViT-B, the RSP-Swin-T-E300 performs better, although the former model has more trainable parameters. It can be observed that the backbones are changing over time. The VGG-16 used in the early years is gradually replaced by the deeper networks such as ResNet-101 or DenseNet-121 due to their better representation ability. 

In the implemented networks, the SeCo-ResNet-50 performs the worst compared with its counterparts, it may be because there still exists a gap between the Sentinel-2 multispectral images where the SeCo trained on with the RGB images for aerial scenes recognition. Compared with the ImageNet pretrained ResNet-50, our RS pretrained ResNet-50 improves the accuracy on all settings. These results imply that RSP brings a better starting point for the optimization of the subsequent finetuning process, attributing to the aerial images used for pretraining compared with the natural images in ImageNet. Similarly, the RSP-Swin-T outperforms IMP-Swin-T on three settings and achieves comparable results on the other two settings. In addition, the ResNet-50 and Swin-T can perform to be competitive compared to other complicated methods by only using the RSP weights without changing the network structures. Besides, when comparing the ImageNet pretrained ResNet-50 and Swin-T, we can find that the IMP-Swin-T performs better in all settings since the vision transformers have stronger context modeling capability. While being initialized by RSP weights, the ResNet becomes to be more competitive and surpasses the IMP-Swin-T on the AID (2:8), NWPU-RESISC (1:9), and NWPU-RESISC (2:8) settings, showing the benefit of RSP again. Owing to the excellent representation ability of ViTAEv2-S, which has both the locality modeling ability and long-range dependency modeling ability, it outperforms both ResNet-50 and Swin-T on almost all the settings, regardless of IMP and RSP. Moreover, the RSP-ViTAEv2-S achieves the best performance compared with all other methods on almost all settings except for the AID (5:5), though on which it also delivers comparable performance with the best one, i.e., RSP-Swin-T-E300. 

In our experiments, RSP helps the networks obtain better performance on small datasets, it may be because the models are easier to converge when adopting the RS pretrained weights. While for the case where training samples are abundant, like AID (5:5), the representation ability of deeper models can be fully exploited. For example, the DenseNet-121 based ESD-MBENeT obtain the best accuracy. Nevertheless, it should be noted that only the feature output from the last layer of RSP-ResNet-50, RSP-Swin-T, or RSP-ViTAEv2-S is used for classification, and it is expected that their performance can be further improved when employing the multilayer intermediate features. In this sense, these RS pretrained models can serve as effective backbones for future research in the aerial recognition field. Furthermore, Table \ref{asr} also shows that the models pretraining with more epochs will probably have stronger representation abilities. Since RSP-ResNet-50-E40 and RSP-Swin-T-E40 fall behind their counterparts with more epochs, we only evaluate the ``E120'' and ``E300'' pretrained weights for these two types of networks in the rest experiments, while for ViTAEv2-S, both the ``E40'' and ``E100'' weights are still used.

\begin{figure}[t]
  \centering
  \includegraphics[width=1\linewidth]{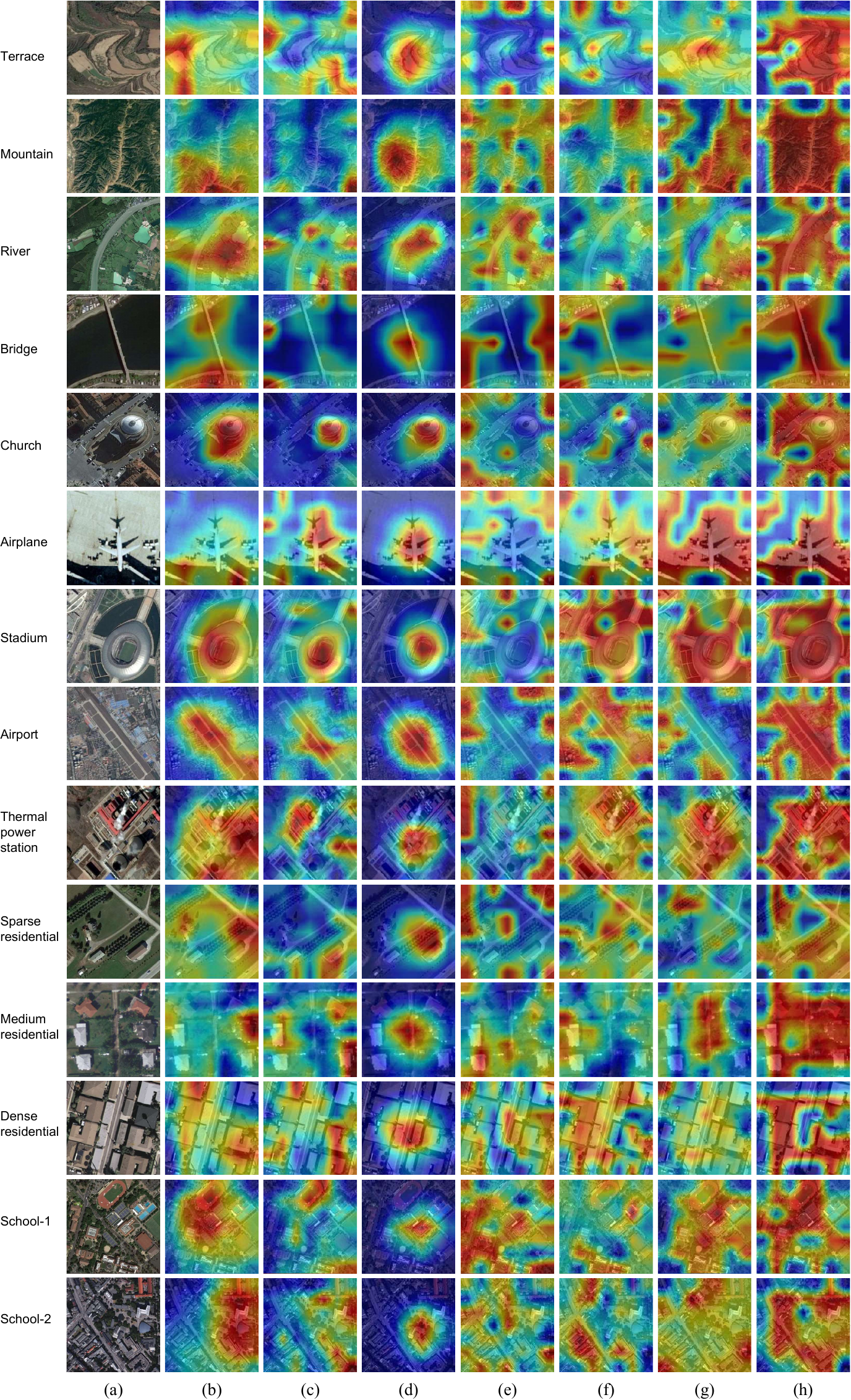}\\
  \caption{Response maps of the evaluated models on different scenes. (a) Original image. (b) IMP-ResNet-50. (c) SeCo-ResNet-50. (d) RSP-ResNet-50. (e) IMP-Swin-T. (f) RSP-Swin-T. (g) IMP-ViTAEv2-S. (h) RSP-ViTAEv2-S.}
  \label{heatmap}
\end{figure}

\textbf{Qualitative Results and Analyses:} Figure \ref{heatmap} shows the response maps of the above evaluated models using Grad-CAM++ \cite{grad_cam_pp} on images from various scenes. The warmer the color is, the higher the response is. To better show the impact of RSP, we use the pretrained weights of ``E300'' for ResNet-50 and Swin-T, and the weights of ``E100'' for ViTAEv2-S. The first three rows are the natural landscapes, and the scenes in 4-8 rows mainly contain specific foreground objects, while the next six rows present some scenes with different artificial constructions. For example, the ``Thermal power station'' scene includes not only chimneys but also cooling towers.

Corresponding to the quantitative results in Table \ref{asr}, the response maps of SeCo-ResNet-50 are scattered and they can not precisely capture the semantic-relevant areas, especially in the natural landscapes or complex scenes with artificial constructions. Compared with the IMP-ResNet-50, the RSP-ResNet-50 pays more attention to the important targets. It implies that RSP facilitates ResNet-50 to learn better semantic representations, probably by seeing semantic-similar images in the MillionAID dataset. Compared to the ResNet-50, the Swin-T has a better context modeling ability by attending faraway regions with the help of MHSA. Thus, their range of high response areas is wider. Surprisingly, the IMP-Swin-T mainly concentrates on background context, but the foreground responses have been enhanced when adopting the RSP. By combining the advantages of CNN and vision transformers, the ViTAEv2-S achieves a comprehensive perception of the whole scene. Especially, the RSP-ViTAEv2-S can better recognize the typical natural RS scenes, such as terrace, mountain and river. In the foreground object based scenes, compared with RSP-ResNet-50, the RSP-ViTAEv2-S not only focuses on the primary objects, but it also considers the related regions in the background. While on the objects, the RSP-ViTAEv2-S assigns higher attention, such as the airplane with a warmer color compared with IMP-ViTAEv2-S. In the residential areas with complex object distributions, the RSP-ViTAEv2-S can correctly capture the sparse buildings and connect these regions to form a holistic representation, effectively perceiving the overall information of the scene. In the first image of the school scene, the RSP-ViTAEv2-S simultaneously focuses on the playground and the surroundings, surpassing the SeCo-ResNet-50. As for the ``school-2'' image that is even difficult to be recognized by humans, these models show different recognition priorities. For example, the RSP-ViTAEv2-S not only considers the campus (it can be possibly distinguished by the irregular shape of buildings) like the IMP-ResNet-50, but also notices the surrounding roads. The results in Table \ref{asr} and Figure \ref{heatmap} validate the effectiveness of RSP and the superiority of vision transformers in the aerial scene recognition task.

\begin{figure}[t]
  \centering
  \subfigure[]{\includegraphics[width=0.48\linewidth]{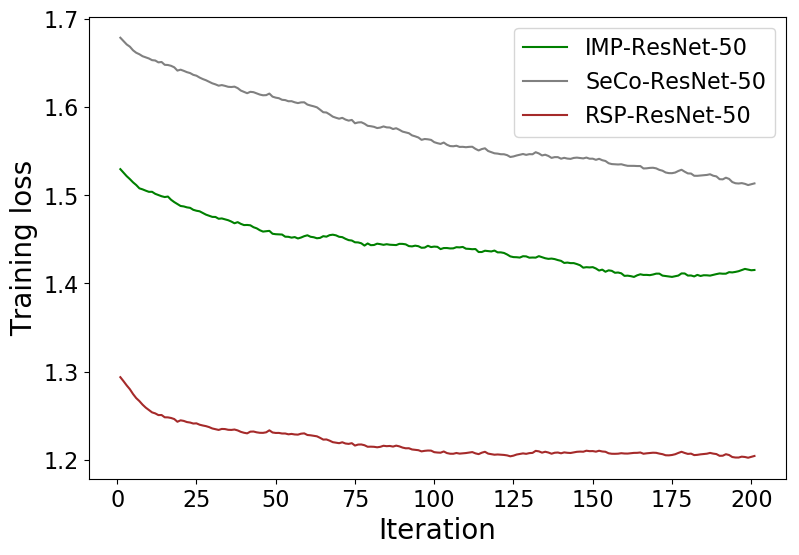}}
  \subfigure[]{\includegraphics[width=0.48\linewidth]{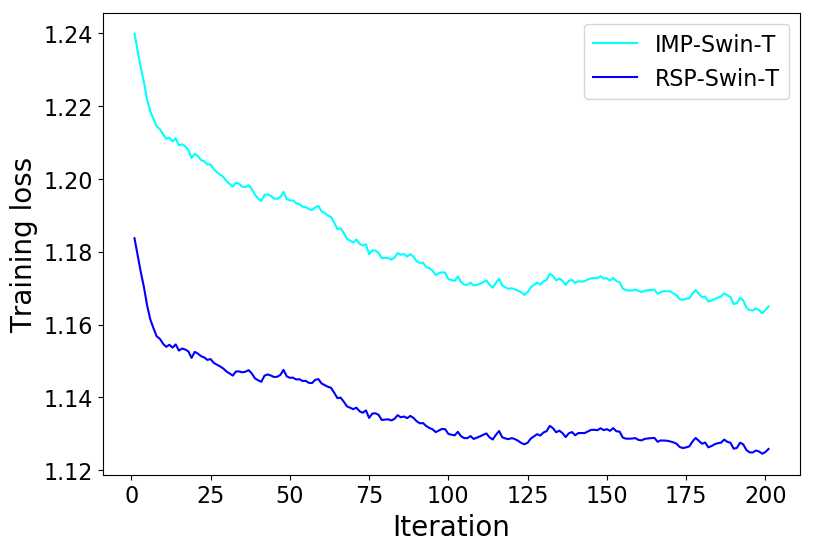}}\\
  \subfigure[]{\includegraphics[width=0.48\linewidth]{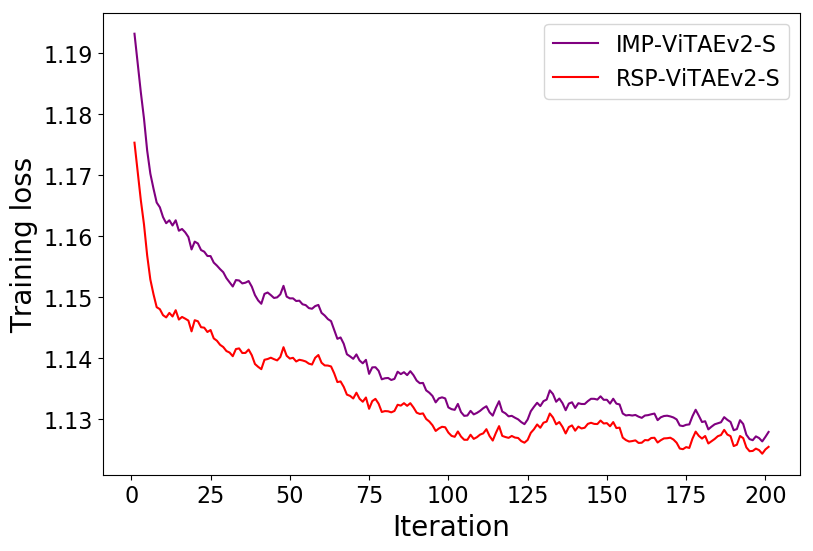}}
  \subfigure[]{\includegraphics[width=0.48\linewidth]{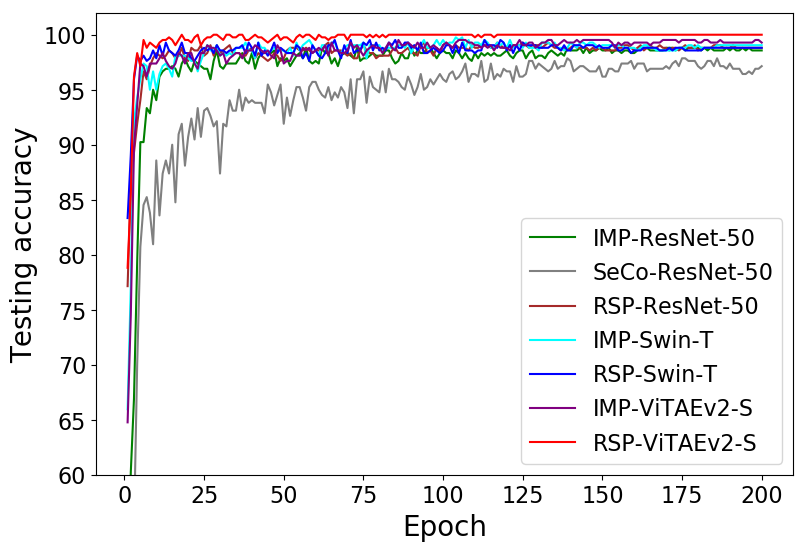}}

  \caption{(a), (b) and (c) are separately the training loss curves of ResNet-50, Swin-T and ViTAEv2-S, where the loss is recorded every 10 iterations. (d) is the testing accuracy curves of these models. All curves are obtained by training in the setting of UCM (8:2). The curves in (a), (b), and (c) have been smoothed by moving average.}
  \label{loss_curve}
\end{figure}

We also provide training loss curves and testing accuracy curves to investigate the performances of different pretraining methods during the training phase. Here, the setting of UCM (8:2) is chosen as our example. The corresponding results have been displayed in Figure \ref{loss_curve}, where the loss curves of three kinds of networks are separately plotted. It can be seen that SeCo-ResNet-50 performs the worst and it has the largest initial loss, further confirming our aforementioned hypothesis that there is a large gap between Sentinel-2 multispectral images and the used RGB aerial images, although they are both RS images. Compared with the IMP, it can be observed that the RS pretrained models have better starting points, proving our intuition that the RS pretrained weights are easier to transfer between aerial scenes. It is also noteworthy that the shapes of these curves are similar for the same network, implying the unique characteristics of different network structures. We can also find the advanced structures enable ViTAEv2-S to reduce the performance gap indicated by the different starting points between IMP and RSP, while other networks failed. In addition, we can also find that the RSP accelerates the learning of ResNet-50 and makes the corresponding accuracy curve similar to Swin-T compared with IMP-ResNet-50. For Swin-T, the RSP also helps it converge fast. When adopting RSP, among all models, the advanced transformer network ViTAEv2-S can simultaneously achieve the best accuracy and the fastest convergence speed.

\subsection{Aerial Semantic Segmentation}

Aerial semantic segmentation is also a classification task like aerial scene recognition but at the pixel-level rather than the scene-level. We then evaluate the above three models on the aerial semantic segmentation task, including the scene parsing and object segmentation subtasks, where the former focuses on labeling each pixel of the whole scene, while the latter emphasizes the segmentation of foreground objects.

\subsubsection{Dataset} 
We use the ISPRS Potsdam dataset\footnote{https://www.isprs.org/education/benchmarks/UrbanSemLab/2d-sem-label-potsdam.aspx} and the large-scale segmentation benchmark --- iSAID \cite{isaid}, to serve as the testbed of the corresponded subtasks, respectively.

\begin{itemize}
  \item Potsdam: This dataset is released by ISPRS Commission WG II/4. It covers a large scene that is collected over 3.42 $\text{km}^2$ area of the Potsdam city. It contains 38 images, whose average size is 6,000 $\times$ 6,000 pixels, and the resolution is 0.5m. Among them, the training and testing sets separately have 24 and 14 images. There are 6 categories included in these scenes, namely impervious surface, building, low vegetation, tree, car, and clutter.
  \item iSAID: This is a large-scale dataset that is mainly for instance segmentation. Also, it provides the semantic mask including 15 foregrounds and 1 background category over the aerial objects. It consists of 2,806 high-resolution images that range from 800 $\times$ 800 to 4,000 $\times$ 13,000 pixels. The training, validation, and test set separately have 1,411, 458, and 937 images. In this paper, only the validation set is used for evaluation since the testing set is unavailable.

\end{itemize}

\begin{figure}[t]
  \centering
  \includegraphics[width=1\linewidth]{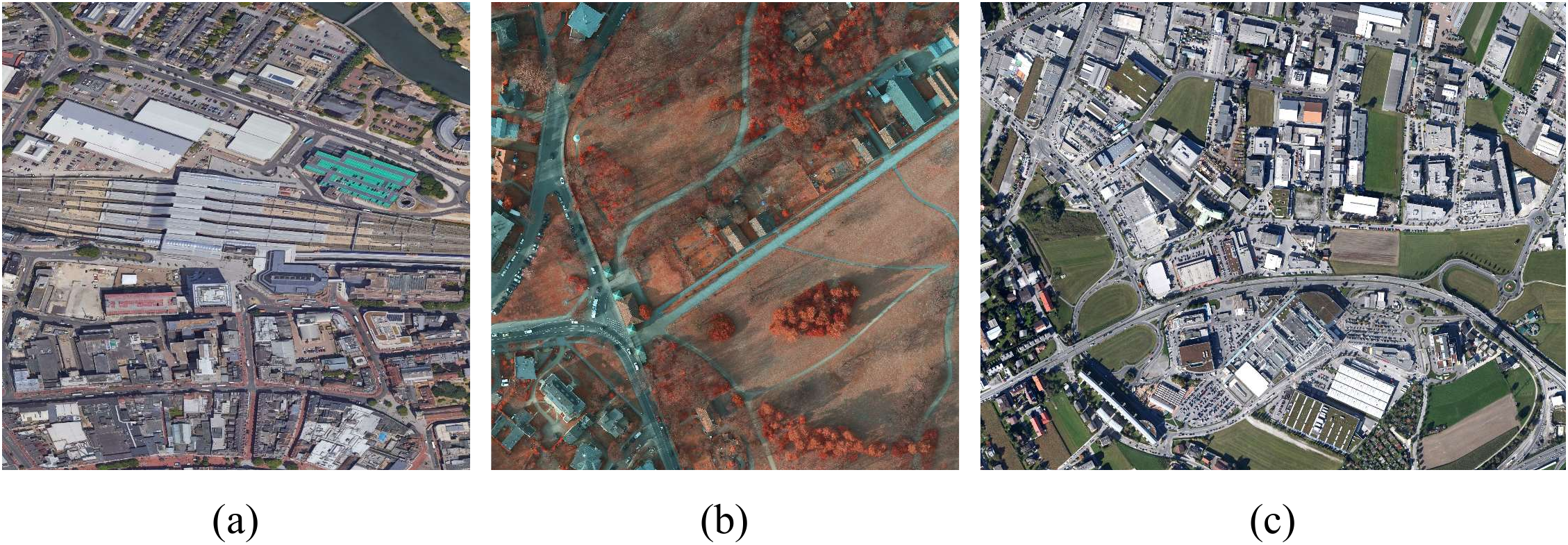}
  \caption{Visual samples from different datasets: (a) MillionAID, (b) Potsdam, and (c) iSAID.}
  \label{seg_datasets}
\end{figure}

\subsubsection{Implementation Detail and Experimental Setting}

We adopt different settings for the above backbone networks by following the common practice. Concretely, the ResNet based networks are trained with the mini-batch stochastic gradient descent with momentum (SGDM) strategy, where the initial learning rate, weight decay, and momentum are separately set to 0.01, 0.0005, and 0.9. The learning rate is optimized by the polynomial scheduler: $current\_lr = min\_lr + initial\_lr \cdot (1-\frac{iter}{max\_iter})^{power}$, where $min\_lr=0.0001, power=0.9$. While the vision transformers, such as the Swin-T and ViTAEv2-S, are trained using the AdamW optimizer, whose learning rate and weight decay are 6e-5 and 0.01. The learning rate schedule adopts the polynomial decay policy with a $power$ of 1.0 and $min\_lr$ of 0. They also have a linear warming up stage at the first 1,500 iterations with the initial learning rate of 1e-6. For a fair comparison, all networks are training 80k iterations with a batch size of 8. Following \cite{swint,xu2021vitae}, we use the UperNet \cite{upernet} as the unified segmentation framework for all the pretrained backbones since the output stride equals 32. All methods are implemented based on mmsegmentation \cite{mmseg2020}. For the convenience of training, the Potsdam and iSAID are separately sampled and cropped into patches with a size of 512 $\times$ 512 and 896 $\times$ 896 with a stride of 384 and 512, respectively. We use the random horizontal flipping data augmentation strategy. Following the evaluation protocol in the aerial segmentation community, for the Potsdam dataset, we report the OA, mean F1 score (mF1), and per class F1 score. Note that the clutter category is regarded as the background and ignored during computing loss and evaluation metrics. While for the iSAID dataset, the intersection over union (IoU) of foreground categories and the average IOU of all classes (including background) are calculated. All evaluations are conducted on a single scale for a fair comparison. 

\subsubsection{Experimental Results}

\begin{table*}[t]
  \scriptsize
  \caption{Results of the UperNet segmentation model with different backbones and SOTA methods on the testing set of the Potsdam dataset. }
  \newcommand{\tabincell}[2]{\begin{tabular}{@{}#1@{}}#2\end{tabular}}
  \centering
  \begin{tabular}{l|l|c|c|c|c|c|c|c}
  \hline
  \multirow{2}*{Method} & \multirow{2}*{Backbone} & \multirow{2}*{OA} & \multirow{2}*{mF1} & \multicolumn{5}{c}{F1 score per category} \\
  \cline{5-9}
  & & & & Imper. surf. & Building & Low veg. & Tree & Car\\
  \hline
  FCN \cite{fcn}& IMP-VGG-16 & 85.59 & 87.61 &  88.61 & 93.29 & 83.29 & 79.83 & 93.02 \\
  S-RA-FCN \cite{ass_2020_tgrs_srafcn}& IMP-VGG-16 & 88.59 & 90.17 & 91.33 & 94.70 & 86.81 & 83.47 & 94.52 \\
  Multi-filter CNN \cite{ass_2018_isprs_mfcnn}& IMP-VGG-16 & 90.65 & 85.23 & 90.94 & 96.98 & 76.32 & 73.37 & 88.55 \\
  FCN \cite{fcn}& IMP-ResNet-50 & 89.42 & 88.66 & 91.46 & 96.63 & 85.99 & 86.94 & 82.28 \\
  DANet \cite{danet}& IMP-ResNet-50 & 89.72 & 89.14 & 91.61 & 96.44 & 86.11 & 88.04 & 83.54 \\
  PSPNet \cite{pspnet}& IMP-ResNet-50 & 89.45 & 90.51 & 91.61 & 96.30 & 86.41 & 86.84 & 91.38 \\
  DeeplabV3+ \cite{deeplabv3_p}& IMP-ResNet-50 & 89.74 & 90.94 & 92.35 & 96.77 & 85.22 & 86.79 & 93.58\\
  ResT \cite{rest}& IMP-ResNet-50 & 89.13 & 90.89 & 91.14 & 95.11 & 86.30 & 87.27 & 94.63 \\
  EaNet \cite{ass_2020_isprs_eanet}&  IMP-ResNet-50 & 90.15 & 91.73 & 92.87 & 96.30 & 86.16 & 87.99 & 95.30\textbf{*} \\
  BotNet \cite{botnet}& IMP-ResNet-50 & 90.42 & 91.77 & 92.34 & 96.30 & 87.32\textbf{*} & 88.74\textbf{*} & 94.17 \\
  LANet \cite{ass_2021_tgrs_lanet} &  IMP-ResNet-50 & 90.84 & 91.95\textbf{*} & 93.05 & 97.19\textbf{*} & 87.30 & 88.04 & 94.19 \\
  \hline
  UperNet & IMP-ResNet-50 & \bfseries 90.64 & \bfseries 89.96 & 92.30 & 96.14 & \bfseries 85.93 & \bfseries 85.66 & 89.76 \\
  UperNet & SeCo-ResNet-50 & 89.64 & 89.03 & 91.21 & 94.92 & 85.12 & 84.89 & 89.02 \\
  UperNet & RSP-ResNet-50 &  90.61   &  89.94 &  \bfseries 92.42  &   \bfseries 96.15    & 85.75 & 85.49 & \bfseries 89.87 \\
  \hline
  UperNet & IMP-Swin-T & \bfseries 91.17 & \bfseries 90.60     &  \bfseries 92.94    & \bfseries 96.66     & \bfseries 86.54 & \bfseries 85.87 & \bfseries 90.98\\
  UperNet & RSP-Swin-T &  90.78 &  90.03     &   92.65    &  96.35     & 86.02 & 85.39 & 89.75  \\
  \hline
  UperNet & IMP-ViTAEv2-S & \bfseries  91.60\textbf{*}    & \bfseries  91.00    & \bfseries  93.34\textbf{*}    & \bfseries  96.84   & \bfseries 87.28 & \bfseries 86.38 & \bfseries 91.18 \\
  UperNet & RSP-ViTAEv2-S &   91.21    &   90.64    &   93.05    &   96.62   &  86.62 & 85.89 & 91.01 \\
  \hline
\end{tabular}
  \label{seg_potsd}
\end{table*}

\begin{table*}[t]
  \scriptsize
  \caption{Results of the UperNet segmentation model with different backbones and SOTA methods on the validation set of the iSAID dataset.}
  \newcommand{\tabincell}[2]{\begin{tabular}{@{}#1@{}}#2\end{tabular}}
  \centering
  \begin{threeparttable}
  \resizebox{\linewidth}{!}{
  \begin{tabular}{l|l|c|c|c|c|c|c|c|c|c|c|c|c|c|c|c|c}
  \hline
  \multirow{2}*{Method} & \multirow{2}*{Backbone} & \multirow{2}*{mIOU} & \multicolumn{15}{c}{IOU per category\tnote{1}}  \\
  \cline{4-18}
  & & & Ship & ST & BD & TC & BC & GTF & Bridge & LV & SV & HC & SP & RA & SBF & Plane & Harbor \\
  \hline
  FCN \cite{fcn} & IMP-VGG-16 & 41.7 & 51.7 & 22.9 & 26.4 & 74.8 & 30.2 & 27.9 & 8.2 & 49.3 & 37.0 & 0 & 30.7 & 51.9 & 52.1 & 62.9 & 42.0 \\
  UNet \cite{unet} & - & 39.2 & 49.0 & 0 & 36.5 & 78.6 & 22.9 & 5.5 & 7.5 & 49.9 & 35.6 & 0 & 38.0 & 46.5 & 9.7 & 74.7 & 45.6 \\
  DenseASPP \cite{denseaspp} & IMP-DenseNet-121 & 56.8 & 61.1 & 50.0 & 67.5 & 86.1 & 56.6 & 52.3 & 29.6 & 57.1 & 38.4 & 0 & 43.3 & 64.8 & 74.1 & 78.1 & 51.1 \\
  DenseUNet \cite{ass_2019_access_denseunet} & IMP-DenseNet-121 & 58.7 & 66.1 & 50.4 & 76.1 & 86.2  & 57.7 & 49.5 & 33.9 & 54.7 & 46.2  & 0 & 45.1 & 65.9 & 71.9 & 82.2 & 54.6 \\
  Semantic FPN \cite{semantic_fpn} & IMP-ResNet-50 & 59.3 & 63.7 & 59.5 & 71.8 & 86.6 & 57.8 & 51.6 & 34.0 & 59.2 & 45.1 & 0 & 46.4 & 68.7 & 73.6 & 80.8 & 51.3 \\
  RefineNet \cite{refinenet} & IMP-ResNet-50 & 60.2 & 63.8 & 58.6 & 72.3 & 85.3 & 61.1 & 52.8 & 32.6 & 58.2 & 42.4 & 23.0 & 43.4 & 65.6 & 74.4 & 79.9 & 51.1 \\
  PSPNet \cite{pspnet} & IMP-ResNet-50 & 60.3 & 65.2 & 52.1 & 75.7 & 85.6 & 61.1 & 60.2 & 32.5 & 58.0 & 43.0 & 10.9 & 46.8 & 68.6 & 71.9 & 79.5 & 54.3 \\
  DeeplabV3 \cite{deeplabv3_arxiv} & IMP-ResNet-50 & 59.0 & 59.7 & 50.5 & 77.0 & 84.2 & 57.9 & 59.6 & 32.9 & 54.8 & 33.7 & 31.3 & 44.7 & 66.0 & 72.1 & 75.8 & 45.7 \\
  DeeplabV3+ \cite{deeplabv3_p} & IMP-ResNet-50 & 60.8 & 63.9 & 52.5 & 72.8 & 84.9 & 56.5 & 58.9 & 32.2 & 59.1 & 42.9 & 31.4 & 46.1 & 67.7 & 72.9 & 79.8 & 52.6 \\
  EMANet \cite{emanet} & IMP-ResNet-50 & 55.4 & 63.1 & 68.4 & 66.2 & 82.7 & 56.0 & 18.8 & 42.1 & 58.2 & 41.0 & 33.4 & 38.9 & 46.9 & 46.4 & 78.5 & 47.5 \\
  ASP-OCNet \cite{ocnet} & IMP-ResNet-50 & 40.2 & 47.3 & 40.2 & 44.4 & 65.0 & 24.1 & 29.9 & 27.1 & 46.3 & 13.6 & 10.3 & 34.6 & 37.9 & 41.4 & 68.1 & 38.0 \\
  DANet \cite{danet} & IMP-ResNet-50 & 57.5 & 60.2 & 63.0 & 71.4 & 84.7 & 50.9 & 52.5 & 28.6 & 57.5 & 42.1 & 30.4 & 46.1 & 40.6 & 63.3 & 80.9 & 48.8 \\
  CCNet \cite{ccnet} & IMP-ResNet-50 & 58.3 & 61.4 & 65.7 & 68.9 & 82.9 & 57.1 & 56.8 & 34.0 & 57.6 & 38.3 & 31.6  & 36.5 &  57.2 &  75.0 & 75.8 & 45.9 \\
  EncNet \cite{encnet} & IMP-ResNet-50 & 58.9 & 59.7 & 64.9 & 70.0 & 84.2 & 55.2 & 46.3 & 36.8 & 57.2 & 38.7 & 34.8 & 42.4 & 59.8 & 69.8 & 76.1 & 48.0 \\
  HRNet \cite{hrnet} & IMP-HRNetW-18 & 61.5 & 65.9 & 68.9 & 74.0 & 86.9 & 59.4 & 61.5 & 33.8 & 62.1 & 46.9 & 14.9 & 44.2 & 52.9 & 75.6 & 81.7 & 52.2 \\
  RANet \cite{asr_2019_cvpr_ranet} & IMP-ResNet-50 & 62.1 & 67.1 & 61.3 & 72.5 & 85.1 & 53.2 & 47.1 & 45.3\textbf{*} & 60.1 & 49.3 & 38.1 & 41.8 & 70.5 & 58.8 & 83.1 & 55.6 \\ 
  AlignSeg \cite{alignseg} & IMP-ResNet-50 & 62.1 & 67.4 & 68.9 & 76.2 & 86.2 & 62.1 & 52.0 & 28.7 & 60.7 & 50.3 & 31.2 & 45.7 & 56.2 & 71.2 & 82.9 & 54.8 \\
  OCR \cite{ocrnet} & IMP-HRNet-W48 & 62.6 & 67.8 & 70.7 & 73.6 & 87.9 & 63.4 & 47.7 & 33.1 & 61.4 & 49.6 & 30.4 & 48.4 & 59.5 & 72.8 & 83.3 & 53.3 \\
  HMANet \cite{hmanet} & IMP-ResNet-50 & 62.6 & 65.4 & 70.9 & 74.7 & 88.7 & 60.5 & 54.6 & 29.0 & 59.7 & 50.3 & 32.6 & 51.4 & 62.9 & 70.2 & 83.8 & 51.9 \\
  FarSeg \cite{farseg} & IMP-ResNet-50 & 63.7 & 65.4 & 61.8 & 77.7 & 86.4 & 62.1 & 56.7 & 36.7 & 60.6 & 46.3 & 35.8 & 51.2 & 71.4\textbf{*} & 72.5 & 82.0 & 53.9 \\
  FactSeg \cite{ass_2022_tgrs_factseg} & IMP-ResNet-50 & 64.8 & 68.3 & 56.8 & 78.4\textbf{*} & 88.9\textbf{*} & 64.9\textbf{*} & 54.6 & 36.3 & 62.7 & 49.5 & 42.7 & 51.5\textbf{*} & 69.4 & 73.6 & 84.1 & 55.7 \\
  \hline
  UperNet & IMP-ResNet-50 & \bfseries 61.9 & \bfseries 65.9 & 73.9 & 68.1 & \bfseries 70.7 & 57.3 & 52.5 & 39.2 & \bfseries 61.2 & \bfseries 48.8 & \bfseries 34.3 & \bfseries 44.5 & 62.1 & \bfseries 76.8 & \bfseries 83.8  & \bfseries 52.2 \\
  UperNet & SeCo-ResNet-50 & 57.2 & 63.9 & 71.7 & 66.9 &69.9  &54.5 & 45.9 &38.9 & 58.2 & 44.8 & 33.2 & 9.3 & 52.3 & 71.6 &83.3 & 51.4 \\
  UperNet & RSP-ResNet-50 & 61.6  & 64.2 & \bfseries 75.9 & \bfseries 68.8 & 69.9 & \bfseries 58.5 & \bfseries 54.4 & \bfseries 40.2 & 59.6 & 47.5 & 32.1 & 43.8 & \bfseries 65.4& 76.5& 82.8 & 51.5 \\
  \hline
  UperNet & IMP-Swin-T & \bfseries 64.6 & \bfseries 69.2 & \bfseries 76.5 & \bfseries 74.1 & 69.9 & 56.3 & 60.1 & 41.9 &\bfseries 62.3& \bfseries 51.6  & \bfseries 44.7\textbf{*} & 45.8 & 64.5 &  75.9 & \bfseries 85.7 & \bfseries 56.7 \\
  UperNet & RSP-Swin-T & 64.1 & 67.0 & 74.6 & 73.7 & \bfseries 70.7 & \bfseries 59.0 & 60.1 & \bfseries 44.3 & 62.0 & 50.6 & 37.6 & \bfseries 46.8 & \bfseries 64.9 & \bfseries 76.2 & 85.2 & 53.8 \\
  \hline
  UperNet & IMP-ViTAEv2-S &\bfseries 65.3\textbf{*} & \bfseries 71.4\textbf{*} & \bfseries 77.5\textbf{*} & 68.2 & \bfseries 71.0 & \bfseries 60.8 & 61.9 & 43.0 & \bfseries 63.8\textbf{*} & \bfseries 53.6\textbf{*} & \bfseries 43.4 & 44.8 & \bfseries 65.1 & \bfseries 77.9\textbf{*} & \bfseries 86.4\textbf{*} & \bfseries 57.7\textbf{*}\\
  UperNet & RSP-ViTAEv2-S & 64.3 & 71.3 & 74.3 & \bfseries 72.2 & 70.4 & 57.4 & \bfseries 63.0\textbf{*} & \bfseries 44.0 & 62.5 & 51.6 & 35.4 & \bfseries 47.0 & 62.2 & 77.7 & 85.2 & 54.7\\
  \hline
\end{tabular}
  }
  \begin{tablenotes}
    \scriptsize
    \item[1] ST: storage tank. BD: baseball diamond. TC: tennis court. BC: baseball court. GTF: ground track field. LV: large vehicle. SV: small vehicle. HC: helicopter.\\ SP: swimming pool. RA: roundabout. SBF: soccer ball field.
  \end{tablenotes}
  \end{threeparttable}
  \label{seg_isaid}
\end{table*}

\begin{figure*}[t]
  \centering
  \includegraphics[width=0.95\linewidth]{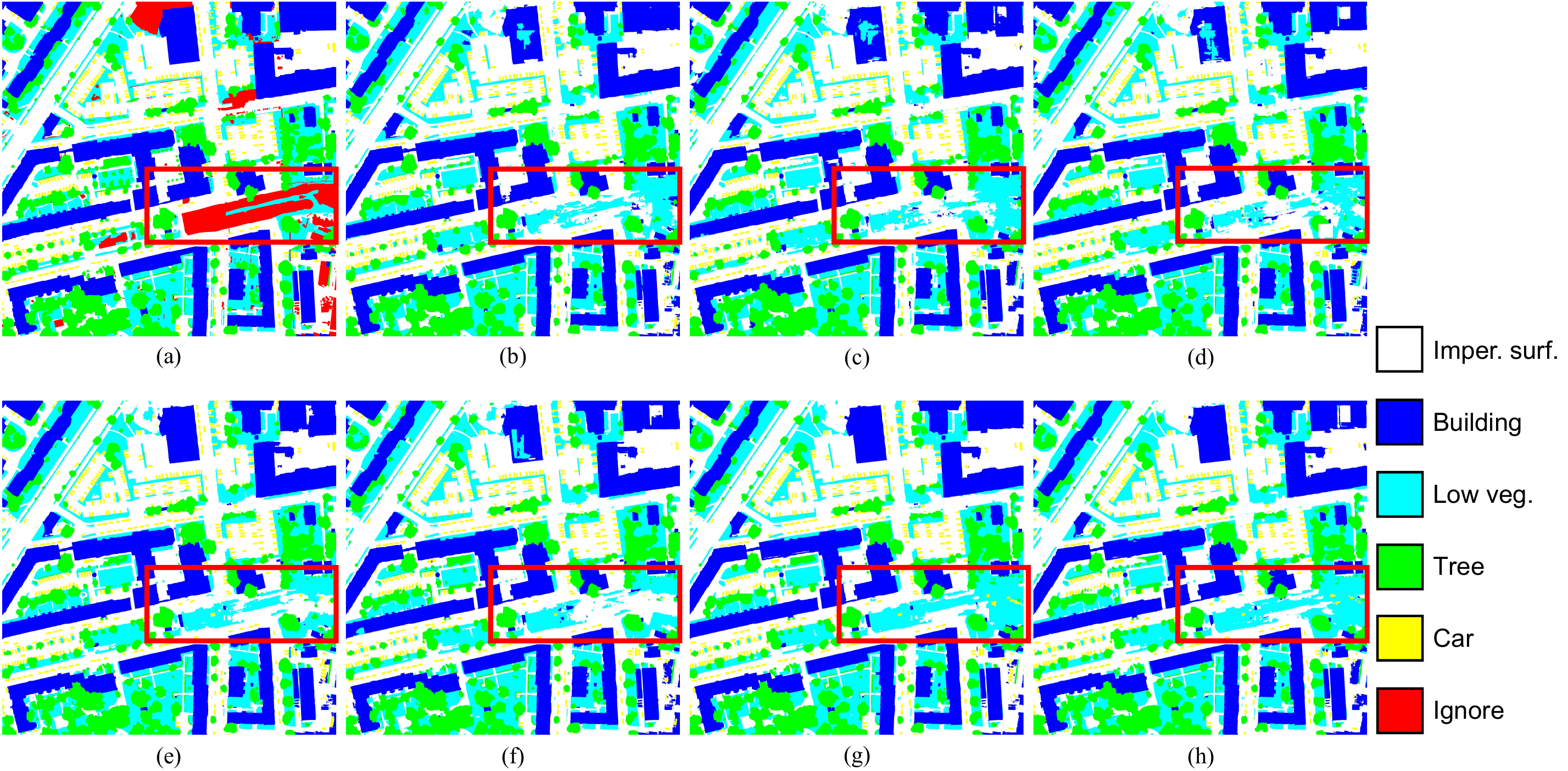}

  \caption{Segmentation maps of the UperNet with different backbones on the Potsdam dataset. (a) Ground Truth. (b) IMP-ResNet-50. (c) SeCo-ResNet-50. (d) RSP-ResNet-50. (e) IMP-Swin-T. (f) RSP-Swin-T. (g) IMP-ViTAEv2-S. (h) RSP-ViTAEv2-S.}
  \label{seg_potsdam_maps}
\end{figure*}

\textbf{Quantitative Results and Analyses:} Table \ref{seg_potsd}-\ref{seg_isaid} present the segmentation results of our methods and other SOTA methods. It can be seen that when changing the backbone from ResNet-50 to Swin-T, and then to ViTAEv2-S, the performance is increased. The results are consistent with the aforementioned scene recognition results, showing the better representation ability of vision transformers. Although the ViTAEv2-S obtains the highest OA on the Potsdam dataset, its mF1 is not as well as LANet \cite{ass_2021_tgrs_lanet}. From Table \ref{seg_isaid}, we can find that the scores of the ``Car'' category of the selected models are worse than other methods. We suspect that it may be because of the encoder-decoder structure and the rough feature fusion strategy in the UperNet, where the high-resolution features have not encoded sufficient high-level semantics, while LANet \cite{ass_2021_tgrs_lanet} not only simultaneously enhance the high and low-level features, it also enriches the semantics of the high-resolution features. Thus, the segmentation performance of the evaluated models based on UperNet on small objects, such as cars, needs to be improved. On the other hand, the IMP-Swin-T performs to be competitive and the IMP-ViTAEv2-S achieves the best performance on the iSAID dataset, outperforming the SOTA methods such as the HRNet \cite{hrnet} and OCR \cite{ocrnet} as well as a series of methods that specially designed for aerial semantic segmentation, e.g., the FarSeg \cite{farseg} and FactSeg \cite{ass_2022_tgrs_factseg}.

Table \ref{seg_isaid} also shows the advantages of RSP models lying in the ``Bridge'' category, which conforms to the finding in the previous scene recognition task. Nevertheless, we can also see from Table \ref{seg_potsd}-\ref{seg_isaid} that, on the segmentation task, the performances of RSP are not as good as the classical IMP. In our considerations, there may be two reasons. The first one is the difference between the pretraining dataset and the evaluation one. Besides the dataset volume (note that the training sample and category numbers of MillionAID are smaller than ImageNet-1k), the spectral disparities also have a side impact on the performance, especially on the Potsdam dataset, which adopts the IR-R-G channels instead of the ordinary RGB image (See Figure \ref{seg_datasets}). Another reason we attribute to the difference between tasks. The representation used for scene recognition needs to have a global understanding of the whole scene as Figure \ref{heatmap} shows, while the segmentation task requires the features to be more detailed while possessing high-level semantic information simultaneously since they separately conduct the scene-level or pixel-level classification. To prove this conjecture, we then evaluate these networks on the aerial object detection task in the next section. The granularity of the representation needed for detection probably lies between those for the segmentation and recognition tasks, since one of the aims in the detection task is the object-level classification.

\textbf{Qualitative Results and Analyses:} We present some visual segmentation results of the UperNet with different backbones on the Potsdam dataset in Figure \ref{seg_potsdam_maps}. As can be seen, only the ViTAEv2-S successfully connects the long strip low vegetations (see the red boxes), while IMP-ViTAEv2-S performs slightly better than RSP-ViTAEv2-S, which is consistent with the quantitative results in Table \ref{seg_potsd}.

\subsection{Aerial Object Detection}

\begin{table*}[ht]
  \scriptsize
  \caption{Results of the ORCN detection model with different backbones and SOTA methods on the testing set of the DOTA dataset. $\dagger$: The result is from AerialDetecton \cite{aerial_det}. $\ddagger$: The result is from the original ORCN paper.}
  \newcommand{\tabincell}[2]{\begin{tabular}{@{}#1@{}}#2\end{tabular}}
  \centering
  \begin{threeparttable}
  \resizebox{\linewidth}{!}{
  \begin{tabular}{l|l|c|c|c|c|c|c|c|c|c|c|c|c|c|c|c|c}
  \hline
  \multirow{2}*{Method} & \multirow{2}*{Backbone} & \multirow{2}*{mAP} & \multicolumn{15}{c}{AP per category\tnote{1}} \\
  \cline{4-18}
  & & & Ship & ST & BD & TC & BC & GTF & Bridge & LV & SV & HC & SP & RA & SBF & Plane & Harbor \\
  \hline
  \bfseries \textit{One-stage} & \multicolumn{17}{c}{}\\
  \hline
  RetinaNet \cite{retinanet} $\dagger$ & IMP-ResNet-50-FPN & 68.43 & 79.11 & 74.32 & 77.62 & 90.29 & 82.18 & 58.17 & 41.81 & 71.64 & 74.58 & 60.64 & 69.67 & 60.60 & 54.75 & 88.67 & 62.57 \\
  DAL \cite{dal} & IMP-ResNet-50-FPN & 71.44 & 79.74 & 78.45 & 76.55 & 90.84 & 79.54 & 66.80 & 45.08 & 76.76 & 67.00 & 60.11 & 73.14 & 62.27 & 57.71 & 88.68 & 69.05  \\
  RSDet \cite{rsdet} & IMP-ResNet-101-FPN & 72.20 & 70.20 & 83.40 & 82.90 & 90.50 & 85.60 & 65.20 & 48.60 & 70.10 & 69.50 & 68.00\textbf{*} & 67.20 & 63.90 & 62.50 & 89.80 & 65.60  \\
  R3Det \cite{r3det} & IMP-ResNet-101-FPN & 71.69 & 77.54 & 83.54 & 81.99 & 90.80 & 81.39 & 62.52 & 48.46 & 74.29 & 70.48 & 60.05 & 67.46 & 59.82 & 61.97 & 89.54 & 65.44 \\
  R3Det \cite{r3det} & IMP-ResNet-152-FPN & 73.74 & 78.21 & 84.23 & 81.17 & 90.81 & 85.26 & 66.10 & 50.53 & 78.66 & 70.92 & 67.17 & 69.83 & 63.77 & 61.81 & 89.49 & 68.16 \\
  S$^2$ANet \cite{aod_2022_tgrs_s2anet} & IMP-ResNet-50-FPN & 74.12 & 87.25 & 85.64 & 82.84 & 90.83 & 84.90 & 71.11 & 48.37 & 78.39 & 78.11  & 57.94 & 69.13 & 62.60 & 60.36 & 89.11 & 65.26 \\
  S$^2$ANet \cite{aod_2022_tgrs_s2anet} & IMP-ResNet-101-FPN & 76.11 & 88.04 & 86.22 & 81.41 & 90.69 & 84.75 & 69.75 & 54.28 & 80.54 & 78.04  & 58.86 & 73.37\textbf{*} & 65.81 & 65.03 & 88.70 & 76.16 \\
  \hline
  \bfseries \textit{Two-stage} & \multicolumn{17}{c}{}\\
  \hline
  ICN \cite{aod_2018_acc_icn} & IMP-ResNet-101-FPN & 68.16 & 69.98 & 78.20 & 74.30 & 90.76 & 79.06 & 70.32 & 47.70 & 67.82 & 64.89 & 50.23 & 64.17 & 62.90 & 53.64 & 81.36 & 67.02 \\
  Faster R-CNN \cite{FasterRCNN} $\dagger$ & IMP-ResNet-50-FPN & 69.05 & 77.11 & 83.90 & 73.06 & 90.84 & 78.94 & 59.09 & 44.86 & 71.49 & 73.25 & 56.18 & 64.91 & 62.95 & 48.59 & 88.44 & 62.18 \\
  CAD-Net \cite{aod_2019_tgrs_cadnet} & IMP-ResNet-101-FPN & 69.90 & 76.60 & 73.30 & 82.40 & 90.90\textbf{*} & 79.20 & 73.50 & 49.40 & 63.50 & 71.10 & 62.20 & 67.00 & 60.90 & 48.40 & 87.80 & 62.00 \\
  ROI Transformer \cite{roi_transformer} & IMP-ResNet-101-FPN & 69.56 & 83.59 & 81.46 & 78.52 & 90.74 & 77.27 & 75.92 & 43.44 & 73.68 & 68.81 & 47.67 & 58.93 & 53.54 & 58.39 & 88.64 & 62.83 \\
  SCRDet \cite{scrdet} & IMP-ResNet-101-FPN & 72.61 & 72.41 & 86.86\textbf{*} & 80.65 & 90.85 & 87.94 & 68.36 & 52.09 & 60.32 & 68.36 & 65.21 & 68.24 & 66.68 & 65.02 & 89.98 & 66.25 \\
  ROI Transformer $\dagger$ \cite{roi_transformer} & IMP-ResNet-50-FPN & 74.61 & 86.87 & 82.51 & 82.60 & 90.71 & 83.83 & 70.87 & 52.53 & 76.67 & 77.93 & 61.03 & 68.75 & 67.61 & 53.95 & 88.65 & 74.67 \\
  Gliding Vertex \cite{glid_vertex} & IMP-ResNet-101-FPN & 75.02 & 86.82 & 86.81 & 85.00 & 90.74 & 79.02 & 77.34\textbf{*} & 52.26 & 73.14 & 73.01 & 57.32 & 70.86 & 70.91\textbf{*} & 59.55 & 89.64 & 72.94 \\
  FAOD \cite{aod_2019_icip_faod} & IMP-ResNet-101-FPN & 73.28 & 79.56 & 84.68 & 79.58 & 90.83 & 83.40 & 76.41 & 45.49 &  68.27 & 73.18 & 64.86 & 69.69 & 65.42 & 53.40 & 90.21\textbf{*} & 74.17 \\
  CenterMap-Net \cite{centermapnet} & IMP-ResNet-50-FPN & 71.74 & 78.10 & 83.61 & 81.24 & 88.83 & 77.80 & 60.65 & 53.15 & 66.55 & 78.62 & 58.70  & 72.36 &  66.19 &  49.36 & 88.88 & 72.10 \\
  FR-Est \cite{aod_2021_tgrs_frest} & IMP-ResNet-101-FPN & 74.20 & 86.44 & 83.56 & 81.17 & 90.82 & 84.13 & 70.19 & 50.44 & 77.98 & 73.52 & 60.55 & 66.72 & 66.59 & 60.64 & 89.63 & 70.59 \\
  Mask OBB \cite{aod_2019_rs_maskobb} & IMP-ResNet-50-FPN & 74.86 & 85.57 & 85.05 & 85.09\textbf{*} & 90.37 & 82.08 & 72.90 & 51.85 & 73.23 & 75.28 & 66.33 & 69.87 & 68.39 & 55.73 & 89.61 & 71.61 \\
  ORCN $\ddagger$ \cite{orcn} & IMP-ResNet-50-FPN & 75.87 & 88.20\textbf{*} & 84.68 & 82.12 & 90.90\textbf{*} & 87.50 & 70.86 & 54.78 & 83.00 & 78.93 & 52.28 & 68.84 & 67.69 & 63.97 & 89.46 & 74.94 \\ 
  ORCN $\ddagger$ \cite{orcn} & IMP-ResNet-101-FPN & 76.28 & 87.52 & 85.33 & 83.48 & 90.90\textbf{*} & 85.56 & 76.92 & 55.27 & 82.10 & 74.27 & 57.28 & 70.15 & 66.82 & 65.51\textbf{*} & 88.86 & 74.36 \\
  \hline
  ORCN & IMP-ResNet-50-FPN & 76.14 & 88.16 & 84.91 & 81.35 & \bfseries 90.90\textbf{*} & \bfseries 87.43 & \bfseries 71.35 & \bfseries 54.86 & \bfseries 83.03 & \bfseries 79.04 & 58.14 & 69.05 & 66.67 & \bfseries 63.39 &89.58  & \bfseries 74.19 \\
  ORCN & SeCo-ResNet-50-FPN & 70.02 & 86.33 & 81.31 & 73.32 &  90.88  & 79.46 & 67.07 & 49.94 & 76.48 & 76.15 & 49.71 & 65.32 & 58.55 & 41.31 &88.64 & 65.90 \\
  ORCN & RSP-ResNet-50-FPN & \bfseries 76.50  & \bfseries 88.17 & \bfseries 85.72 & \bfseries 81.88 & 90.84 & 86.17 & 70.91 & 54.39 & 83.01 & 78.67 & \bfseries 62.22 & \bfseries 72.21 &  \bfseries 67.45& 62.22& \bfseries 89.78 & 73.99 \\
  \hline
  ORCN & IMP-Swin-T-FPN & 76.07 & \bfseries 88.02 & \bfseries 84.92 & \bfseries 82.23 & \bfseries 90.90\textbf{*} & \bfseries 87.42 & 74.37 & 52.25 & 83.55 & 77.99 & \bfseries 63.07 & 69.30 & 65.99 & 57.70 & 89.48 & 73.88 \\
  ORCN & RSP-Swin-T-FPN & \bfseries 76.12 & 87.83 & 84.84 & 79.74 & 90.86 & 85.90 & \bfseries 74.50 & \bfseries 52.91 & \bfseries 84.02 & \bfseries 78.96 & 57.36 & \bfseries 70.61 &  \bfseries 67.33 & \bfseries 62.90 & \bfseries 89.54 & \bfseries 74.45 \\
  \hline
  ORCN & IMP-ViTAEv2-S-FPN & 77.38 & \bfseries 88.14 & \bfseries 86.35 & \bfseries 83.50 & 90.90\textbf{*} & 87.51 & \bfseries 75.38 & 53.42 & \bfseries 85.15\textbf{*} & \bfseries 79.99\textbf{*} & 66.03 & 66.12 & \bfseries 70.91\textbf{*} & 61.02 & 89.27 & \bfseries 76.95\textbf{*}  \\
  ORCN & RSP-ViTAEv2-S-FPN & \bfseries 77.72\textbf{*} & 88.04 & 85.58 & 83.04 & 90.90\textbf{*} & \bfseries 88.17\textbf{*} & 75.16 & \bfseries 55.85\textbf{*} & 84.34 & 79.95 & \bfseries 67.89 & \bfseries 67.15 & 70.60 & \bfseries 62.64 & \bfseries 89.66 & 76.77 \\
  \hline
\end{tabular}
  }
  \begin{tablenotes}
    \scriptsize
    \item[1] ST: storage tank. BD: baseball diamond. TC: tennis court. BC: baseball court. GTF: ground track field. LV: large vehicle. SV: small vehicle. HC: helicopter.\\ SP: swimming pool. RA: roundabout. SBF: soccer ball field.
  \end{tablenotes}
  \end{threeparttable}
  \label{det_dota}
\end{table*}

\begin{table}[h]
  \scriptsize
  \caption{Results of the ORCN detection model with different backbones and SOTA methods on the testing set of the HRSC2016 dataset. $\dagger$: The result is from the original ORCN paper.}
  \newcommand{\tabincell}[2]{\begin{tabular}{@{}#1@{}}#2\end{tabular}}
  \centering
  \begin{tabular}{l|l|c}
  \hline
  Method & Backbone & mAP \\
  \hline
  R2PN \cite{aod_2018_grsl_r2pn} & IMP-VGG-16 & 79.6 \\
  RRD \cite{rrd} & IMP-VGG-16 & 84.3 \\
  FoRDet \cite{aod_2022_tgrs_fordet} & IMP-VGG-16 & 89.9\\
  R2CNN \cite{r2cnn} & IMP-ResNet-101 & 73.1 \\
  Rotated RPN \cite{aod_2018_tmm_rotated_rpn} & IMP-ResNet-101 & 79.1  \\
  ROI Transformer \cite{roi_transformer} & IMP-ResNet-101-FPN & 86.2  \\
  Gliding Vertex \cite{glid_vertex} & IMP-ResNet-101-FPN & 88.2 \\
  GRS-Det \cite{aod_2021_tgrs_grsdet} & IMP-ResNet-50-FPN & 88.9 \\
  GRS-Det \cite{aod_2021_tgrs_grsdet} & IMP-ResNet-101-FPN & 89.6\\
  R3Det \cite{r3det} & IMP-ResNet-101-FPN & 89.3  \\
  DAL \cite{dal} & IMP-ResNet-101-FPN & 89.8 \\
  AproNet \cite{aod_2022_tgrs_s2anet} & IMP-ResNet-101-FPN & 90.0 \\
  S$^2$ANet \cite{aod_2022_tgrs_s2anet} & IMP-ResNet-101-FPN & 90.2 \\
  ORCN $\dagger$ \cite{orcn} & IMP-ResNet-50-FPN & 90.4 \\
  CHPDet \cite{aod_2022_tgrs_chpdet} & Hourglass104 & 90.6\textbf{*} \\
  \hline
  ORCN & IMP-ResNet-50-FPN & \bfseries 90.4\\
  ORCN & SeCo-ResNet-50-FPN & 88.9 \\
  ORCN & RSP-ResNet-50-FPN & 90.3 \\
  \hline
  ORCN & IMP-Swin-T-FPN & 89.7 \\
  ORCN & RSP-Swin-T-FPN & \bfseries 90.0 \\
  \hline
  ORCN & IMP-ViTAEv2-S-FPN & 90.4 \\
  ORCN & RSP-ViTAEv2-S-FPN & 90.4 \\
  \hline
\end{tabular}
  \label{det_hrsc}
\end{table}

Since the aerial images are top-down photoed in the sky, the objects can be presented in any direction in the birdview. Thus, the aerial object detection is the oriented bounding box (OBB) detection, which is distinguished from the usual horizontal bounding box (HBB) task on natural images \cite{FasterRCNN,retinanet,zj_detection4}. In this paper, similar to segmentation, we also use different detection datasets in the experiments. Concretely, we evaluated on the multi-category RS objects detection and the single-category ship detection subtasks, respectively.

\subsubsection{Dataset}
Two datasets including the large-scale DOTA \cite{dota1} scenes and the commonly used HRSC2016 \cite{hrsc2016} dataset are separately utilized for the above objectives.

\begin{itemize}
  
  \item DOTA: This is the most famous large-scale dataset for OBB detection. It totally contains 2,806 images whose size ranges from 800 $\times$ 800 to 4,000 $\times$ 4,000, where 188,282 instances belonging to 15 categories are included. The training, validation, and testing set separately have 1,411/458/937 tiles. It should be noticed that the categories are completely the same with the iSAID dataset, since the two datasets share the same set of scenes. The difference lies in the annotations for different tasks.
  \item HRSC2016: This is a specialized ship detection dataset, where the bounding boxes are annotated in arbitrary orientations. 1,061 images with the size ranging from 300 $\times$ 300 to 1,500 $\times$ 900 are included. In the official division, 436/181/444 images are used for training, validation, and testing, respectively. The dataset only has one category, since there is no need to recognize the type of ships.

\end{itemize}

\subsubsection{Implementation Detail and Experimental Setting}

Similar to segmentation, the ResNet models are trained using the SGDM algorithm with a learning rate of 0.005, a momentum of 0.9, and a weight decay of 0.0001, while the vision transformers are trained with the AdamW optimizer, where the learning rate and weight decay are separately set to 0.0001 and 0.05. These models are trained for 12 and 36 epochs with a batch size of 2 on DOTA and HRSC2016 scenes, respectively. The learning rate is adjusted by a multi-step scheduler. On the DOTA dataset, the learning rate will be separately reduced by 10$\times$ after the 8th epoch and the 11th epoch, while on the HRSC2016 scene, the corresponded settings are epoch 24 and epoch 33. We use one of the SOTA OBB detection frameworks --- ORCN \cite{orcn} to evaluate the performance of different pretrained backbones. We adopt the default hyper-parameters of ORCN, which is implemented in OBBDetection\footnote{https://github.com/jbwang1997/OBBDetection}. Following \cite{orcn}, the DOTA dataset is sampled and cropped to 1,024 $\times$ 1,024 patches with a stride of 824, while the HRSC2016 images are scaled keeping the aspect ratio with the shorter side equals to 800, and the length of the longer side is less than or equal to 1333. Data augmentations during training include random horizontal and vertical flipping. For convenience, the original training and validation sets are merged for training, while the original testing sets of DOTA and HRSC2016 are separately used for evaluation. We report the mean average precision (mAP) of all categories and the average precision (AP) of each class on the corresponding testing set. All models are trained on a single V100 GPU.

\subsubsection{Experimental Results}

\textbf{Quantitative Results and Analyses:} Table \ref{det_dota}-\ref{det_hrsc} show the results of OBB detection experiments. On the challenging DOTA dataset, it can be seen that using the advanced ORCN framework, the models whose backbone is either ResNet-50 or Swin-T performs well, although the mAPs of Swin-T models are slightly lower than the ResNet models. The ViTAEv2-S, which is a kind of vision transformer network that is introduced the inductive biases including the locality and scale-invariance characteristics of CNN, obtains amazing performance that improves the ORCN baseline by nearly 2\% mAP. Another point needed to be noticed is the performance of RSP weights on these three backbones all outperforms their ImageNet pretrained counterparts. These results support our previous claims that the granularity of the representation required for the detection task is closer to that for the scene recognition task compared with the segmentation task. Thus, the performance difference between RSP and IMP in the detection experiments aligns with the results in the scene recognition experiments. 

In addition, we observe that compared with IMP-ViTAEv2-S, the APs of most categories obtained by RSP-ViTAEv2-S are smaller, implying the universality of IMP. Nevertheless, the mAP of RSP-ViTAEv2-S is higher than IMP-ViTAEv2-S, since RSP has significant advantages in the categories of ``Bridge'' and aerial vehicles including ``Helicopter'' or ``Plane'', echoing the previous finding in the segmentation experiments. While on the other categories, the gaps between these two models are not very large. Combining the above two points, RSP-ViTAEv2-S delivers better overall performance than IMP-ViTAEv2-S. On HRSC2016 dataset, the CHPDet \cite{aod_2022_tgrs_chpdet} that performs the best is a specifically designed detector by considering the ship characteristics. For ORCN \cite{orcn} related networks, the results of RSP and IMP are roughly the same, where there are wins or losses on both sides. Compared with CNN, the vision transformer models have not demonstrated the advantages. We think that on this relatively easy subtask, where only one category needed to be detected and the ship sizes in HRSC2016 are relatively larger than DOTA, the performance is probably saturated. 

\begin{figure}[t]
  \centering
  \includegraphics[width=1\linewidth]{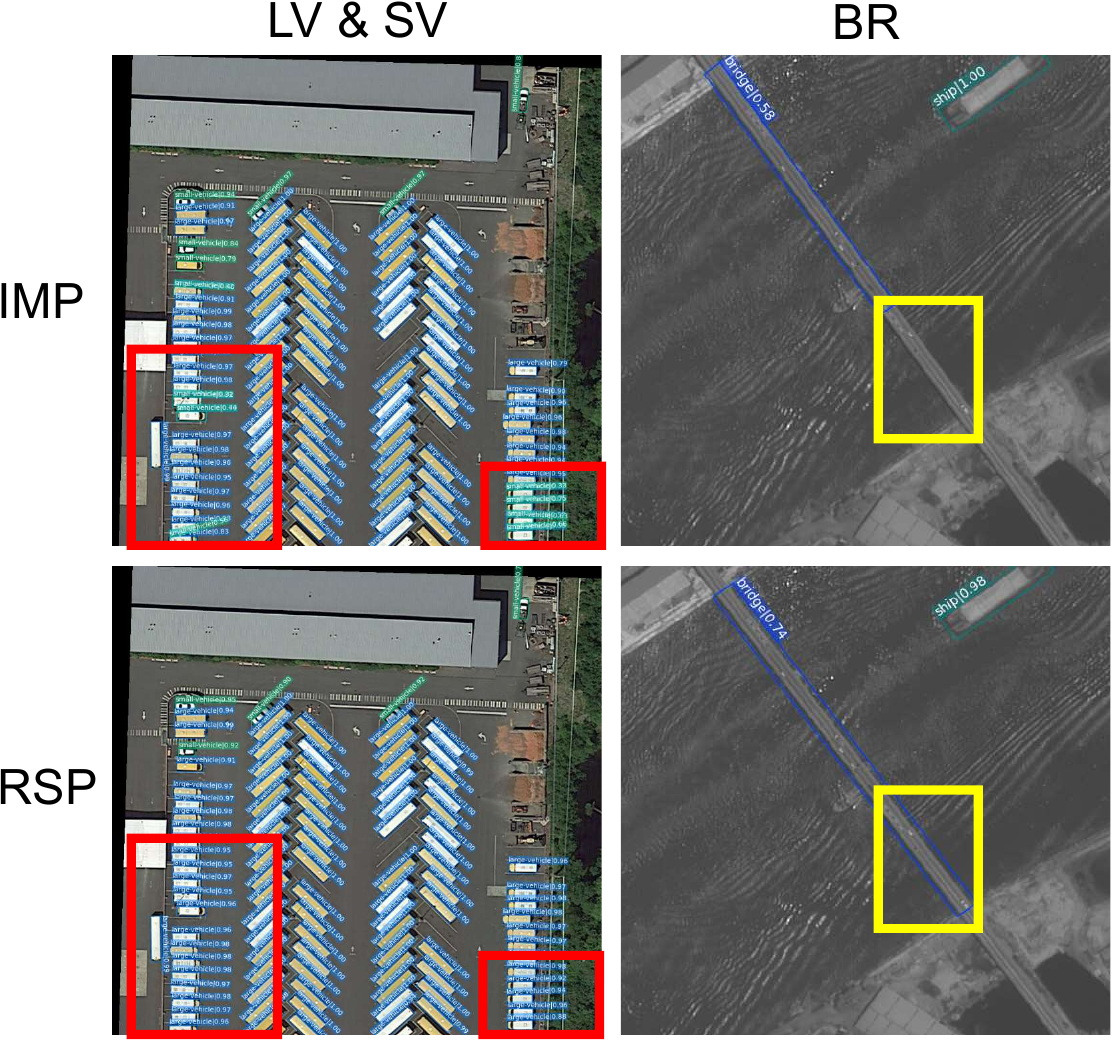}
  \caption{Visual detection results of the ORCN model with the ViTAEv2-S backbones on the DOTA testing set. LV: large vehicle. SV: small vehicle. BR: Bridge. IMP: IMP-ViTAEv2-S. RSP: RSP-ViTAEv2-S.}
  \label{det_maps}
\end{figure}

\textbf{Qualitative Results and Analyses:} We visualize some detection results of the ORCN model with the ViTAEv2-S backbones on the DOTA testing set in Figure \ref{det_maps}. The red boxes show that, when objects are densely distributed, the RSP-ViTAEv2-S can still predict correct object categories, while the IMP-ViTAEv2-S is confused by the dense context and makes wrong predictions. For the ``Bridge'' category, the IMP-ViTAEv2-S produces missing detections (see yellow boxes), while the RSP-ViTAEv2-S model successfully detects the long bridge with a high confidence score.

\subsection{Aerial Change Detection}

We then apply the above models on a typical application in the RS field, i.e., change detection, which aims to find the changes between two aerial images of a same region captured at different times. It is formulated as a pixel-level binary classification task, where ``1'' indicates change.

\subsubsection{Dataset}

We adopt the commonly used CDD \cite{cdd} and LEVIR \cite{levir} datasets to comprehensively evaluate the above models on this task since they separately involve the natural and artificial changes.

\begin{itemize}
  \item CDD: The original dataset contains 11 pairs of multi-source real season-varying RS images collecting from GE, where 7 pairs of images are at the size of 4,725 $\times$ 2,200 and 4 image pairs are at the size of 1,900 $\times$ 1,000 pixels. The resolutions are ranging from 0.03m to 1m. Then, \cite{cdd_clip} clipped the images to a series of 256 $\times$ 256 patches and generated a dataset, where the sizes of the training, validation and testing set are 10,000/3,000/3,000, respectively.
  \item LEVIR: This dataset is collected using the GE API on 20 different regions in Texas, the USA, from 2002 to 2018. It contains 637 image pairs at the size of 1,024 $\times$ 1,024 and with a high resolution of 0.5m, where most changes are from man-made structures, including 31,333 independent building change entities. The training, validation, and testing set contain 445/64/128 image pairs, respectively.

\end{itemize}

\subsubsection{Implementation Detail and Experimental Setting}

In this section, we adopt a SOTA framework --- BIT \cite{acd_2021_tgrs_bit}, which uses the transformer to capture the contextual information between different temporal images for change detection. If BIT is equipped with the ResNet backbone, it is optimized by the SGDM optimizer, where the learning rate, momentum, and weight decay are 0.001, 0.99, and 0.0005. While the Swin or ViTAE based BIT models are trained with the AdamW optimizer with the learning rate of 6e-5 and weight decay of 0.01. These models are trained for 200 epochs with a batch size of 8, while the learning rate is linearly decayed until the end of training. Following \cite{acd_2021_tgrs_bit}, the LEVIR dataset is clipped to the patches at the size of 256 $\times$ 256 with no overlaps. Thus, the sizes of the training, validation and testing set are 7,120/1,024/2,048. The final performance of different models is evaluated on the testing set, while the results on the validation set are only used to select the best model during training. We use the F1 score as the evaluation metric and the experiments are conducted on a single V100 GPU.

\subsubsection{Experimental Results}

\begin{table}[t]
  \scriptsize
  \caption{Results of the BIT change detection model with different backbones and SOTA methods on the testing set of CDD and LEVIR datasets. $\dagger$: The result is from the original BIT paper.}
  \newcommand{\tabincell}[2]{\begin{tabular}{@{}#1@{}}#2\end{tabular}}
  \centering
  \begin{tabular}{l|l|c|c}
  \hline
  \multirow{2}*{Method} & \multirow{2}*{Backbone} & \multicolumn{2}{c}{F1 score} \\
  \cline{3-4}
  & & CDD & LEVIR \\
  \hline
  FC-EF \cite{fcsn_cd} & --- & 77.11  & 62.32 \\
  FC-Siam-conc \cite{fcsn_cd} & --- & 82.50 & 68.21 \\
  FC-Siam-diff \cite{fcsn_cd}& --- & 83.73 & 63.09 \\
  CLNet \cite{acd_2021_isprs_clnet}& --- & 92.10 & 90.00 \\
  SNUNet-c48 \cite{acd_2021_grsl_snunet}& --- & 96.20 & --- \\
  IFN \cite{acd_2020_isprs_ifn}& IMP-VGG-16 & 90.30 & 83.57 \\
  DASNet \cite{acd_2021_jstars_dasnet}& IMP-VGG-16 & 91.93 & 82.83 \\
  SRCDNet \cite{acd_2021_tgrs_srcdnet}& IMP-ResNet-18 & 90.02 & --- \\
  STANet \cite{levir}& IMP-ResNet-18 & 90.75 & 87.34 \\
  BSFNet \cite{acd_2021_grsl_bsfnet}& IMP-ResNet-18 & 91.90 & 88.00\\
  DSAMNet \cite{acd_2021_tgrs_dsamnet}& IMP-ResNet-18 & 93.69 & --- \\
  HRTNet \cite{acd_2021_isprs_hrtnet}& IMP-HRNet-W18 & 93.71 & 88.48 \\
  CDNet+IAug \cite{acd_2021_tgrs_cdnet_iaug}& IMP-ResNet-18 & --- & 89.00 \\
 BIT$\dagger$ \cite{acd_2021_tgrs_bit}& IMP-ResNet-18 & --- & 89.31 \\
 ChangeFormer \cite{changeformer}& IMP-MiT-B2 \cite{segformer} & --- & 90.40 \\
  CS-HSNet \cite{acd_2021_jstars_cshsnet}& IMP-ResNet-50 & 94.95 & 90.79 \\
 LSS-Net \cite{acd_2021_jstars_lssnet}& IMP-SE-ResNet-50 & 96.30 & --- \\
 ChangeStar \cite{changestar} & IMP-ResNext-101-32$\times$4d & --- & 91.25 \\
  \hline
  BIT  & IMP-ResNet-50 & 95.09 & 89.19\\
  BIT  & SeCo-ResNet-50 & 95.95 &\bfseries 90.14 \\
  BIT  & RSP-ResNet-50 &\bfseries 96.00 & 90.10 \\
  \hline
  BIT  & IMP-Swin-T & 94.77 &\bfseries 90.25 \\
  BIT  & RSP-Swin-T &\bfseries 95.21 & 90.10 \\
  \hline
  BIT  & IMP-ViTAEv2-S & \bfseries 97.02\textbf{*} & \bfseries 91.26\textbf{*} \\
  BIT  & RSP-ViTAEv2-S & 96.81 & 90.93 \\
  \hline
\end{tabular}
  \label{cd_cdd_levir}
\end{table}

\begin{figure*}[t]
  \centering
  \includegraphics[width=1\linewidth]{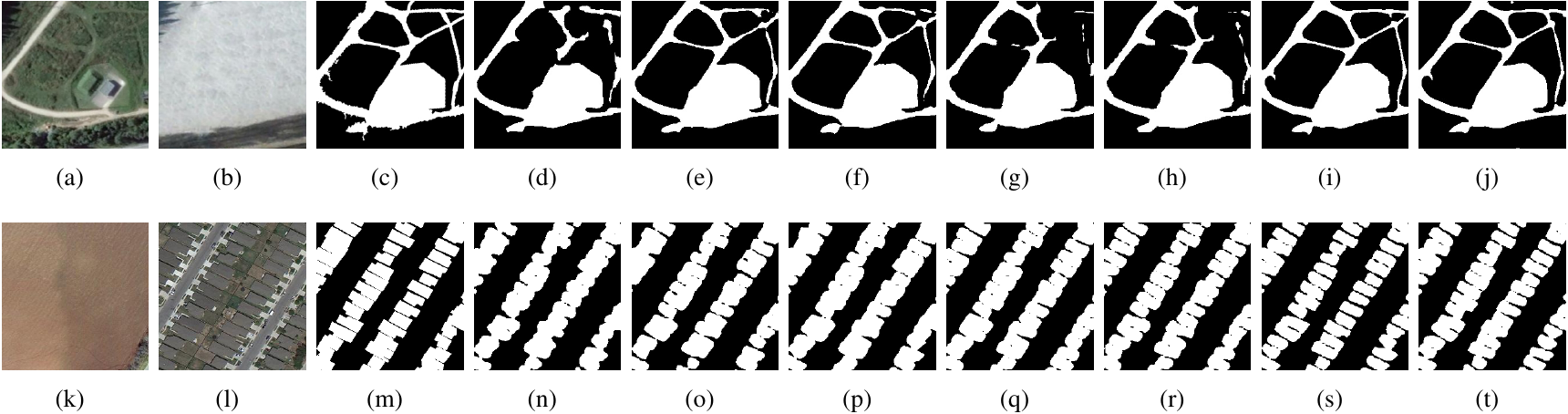}

  \caption{Visual change detection results. The first and second row separately show the change detection results of a sample image from the CDD and LEVIR datasets. Here, (a) and (k), (b) and (l) are the first and second temporals of the same regions. (c) and (m) are ground truth change annotations. (d) and (n) are the results of the IMP-ResNet-50 based BIT, while (e) and (o), (f) and (p), (f) and (o), (g) and (q), (h) and (r), (i) and (s), (g) and (t) are the results from the SeCo-ResNet-50, RSP-ResNet-50, IMP-Swin-T, RSP-Swin-T, IMP-ViTAEv2-S, and RSP-ViTAEv2-S backbones, respectively.}
  \label{cd_maps}
\end{figure*}

\textbf{Quantitative Results and Analyses:} The quantitative results are summarized in Table \ref{cd_cdd_levir}. Without surprise, the self-supervised SeCo pretrained weights perform well on this task, e.g., the SeCo-ResNet-50 based BIT performs better than the IMP counterpart. Although the SeCo weights are trained to achieve seasonal invariance, the change features can be encoded via the multi-head sub-space embedding \cite{seco}. Nevertheless, ViTAEv2-S pretrained either by IMP or RSP performs better than SeCo-ResNet-50, showing the benefit of using the advanced backbone.

Compared with other methods, it is no doubt that the ViTAEv2-S achieves the best performance, showing the potentiality of applying an advanced vision transformer on RS field. As before, we analyze the performance difference between the RSP with the IMP through the perspective of task characteristics. Given the analyses in previous sections, we can infer that the granularity of the required representation for the change detection lies in between those of the segmentation and detection, since it is also a segmentation task, although there only are two categories and there is no need to recognize specific object category. 

\textbf{Qualitative Results and Analyses:} We present some visual change detection results in Figure \ref{cd_maps}. As can be seen, ResNet-50 and Swin-T by IMP can not well detect the changes in roads inside the fields in the natural scene. This issue could be partly alleviated by adopting RSP. It is consistent with the results in Table \ref{cd_cdd_levir} that SeCo-ResNet-50 further improves the detection in the road areas. Compared with the above models, the ViTAEv2-S model effectively captures the road details. In the artificially changed scene, the ViTAEv2-S model greatly overcomes the problem of object adhesion in the results of all other models, demonstrating that the ViTAEv2-S features are more discriminative for distinguishing objects from the background.

\begin{table}[t]
  \scriptsize
  \caption{Overall comparisons of different backbones trained with different epochs on all downstream tasks.}
  \newcommand{\tabincell}[2]{\begin{tabular}{@{}#1@{}}#2\end{tabular}}
  \centering
  \resizebox{\linewidth}{!}{
  \begin{tabular}{l|c|c|c|c}
  \hline
  Backbone & \tabincell{c}{Scene \\ Recognition} & \tabincell{c}{Semantic\\ Segmentation} & \tabincell{c}{Object\\ Detection} & \tabincell{c}{Change\\ Detection} \\
  \hline
  RSP-ResNet-50-E120 & 96.53 & \bfseries 75.70  & 82.51 & 92.82 \\
  RSP-ResNet-50-E300 & \bfseries 96.63 & 75.68  & \bfseries 83.39 & \bfseries 93.05 \\
  \hline
  RSP-Swin-T-E120 & 96.06 & 76.17  & \bfseries 83.08 & 92.28 \\
  RSP-Swin-T-E300 & \bfseries 96.44 & \bfseries 77.07  & 82.80 & \bfseries 92.66 \\
  \hline
  RSP-ViTAEv2-S-E40 & 96.76 & 76.95  & 83.65 & 93.08 \\
  RSP-ViTAEv2-S-E100 & \bfseries 97.01 & \bfseries 77.45  & \bfseries 83.98 & \bfseries 93.87 \\
  \hline
\end{tabular}
  }
  \label{different_epoch}
\end{table}

\subsection{Overall Comparison of Different Backbones on All Tasks}
In this part, we comprehensively compare the performance of different backbones by RSP on all tasks. Specifically, the scores in Table \ref{different_epoch} are calculated by averaging the scores across all datasets for each task. For example, we calculate the average score of the mean values on five settings in the scene recognition task, so the overall accuracy of RSP-ResNet-50-E120 is $(99.52+96.60+97.78+93.76+94.97)/5 \approx 96.53$, while the overall score of RSP-Swin-T-E300 on the segmentation task is obtained by averaging the mF1 on Potsdam dataset and the mIOU on iSAID dataset: $(90.03+64.10)/2 \approx 77.07$. From Table \ref{different_epoch}, we can find that the backbones pretraining with more epochs generally perform better on downstream tasks, since they obtain stronger representations, although there is an exception, i.e., the Swin-T on the object detection task, implying the task discrepancy also matters. In general, ViTAEv2-S takes advantage of CNN and transformers and delivers better performance when training with more epochs, and outperforms ResNet-50 and Swin-T on all the tasks.

\section{Conclusion}

In this study, we investigate the remote sensing pretraining problem based on both CNN and vision transformers on the largest remote sensing dataset MillionAID, and comprehensively evaluate their performance on four related tasks, including scene recognition, semantic segmentation, object detection, and change detection, as well as compare them with the ImageNet pretraining and other SOTA methods. By synthetically analyzing the experiment results, we draw the following conclusions:

\begin{itemize}
  \item Compared with the traditional CNN model, the vision transformers perform competitively on a series of remote sensing tasks, and they can obtain better performance on some more challenging datasets, such as the iSAID and DOTA. Particularly, the ViTAEv2-S, an advanced model introducing the inductive biases of CNN into the vision transformers, achieves the best performance on almost all settings of these tasks.
  \item Benefitting from the large capacity of ImageNet-1K dataset, the classical IMP enables deep models to learn more universal representations that generalize well to almost all categories in the downstream tasks. Thus, the IMP can produce competitive baseline results despite aerial scenes. RSP is comparable with IMP and performs extremely well on some specific categories, such as the ``Bridge'' and ``Airplane'', owing to mitigating the data-level discrepancy between the upstream pretraining task and downstream task.
  \item The task-level discrepancy also has a side impact on the performance of RSP. If the granularity of the representation required for a specific downstream task is closer to that of the upstream pretraining task, i.e., scene recognition, RSP usually leads to better performances.

\end{itemize}

We hope this study can provide useful insights to the community about using advanced vision transformers and remote sensing pretraining. In future work, we will investigate the RSP on large-scale datasets for downstream tasks as well as the unsupervised pretraining considering the abundant unlabelled data in this area.

\ifCLASSOPTIONcaptionsoff
  \newpage
\fi



\bibliographystyle{IEEEtran}
\bibliography{RSP}

\end{document}